\documentclass[runningheads]{llncs}

\usepackage{eccv}

\definecolor{turquoise}{cmyk}{0.65,0,0.1,0.3}
\definecolor{purple}{rgb}{0.65,0,0.65}
\definecolor{dark_green}{rgb}{0, 0.5, 0}
\definecolor{orange}{rgb}{0.8, 0.6, 0.2}
\definecolor{red}{rgb}{0.8, 0.2, 0.2}
\definecolor{darkred}{rgb}{0.6, 0.1, 0.05}
\definecolor{blueish}{rgb}{0.3, 0.3, .6}
\definecolor{light_gray}{rgb}{0.7, 0.7, .7}
\definecolor{pink}{rgb}{1, 0, 1}
\definecolor{greyblue}{rgb}{0.25, 0.25, 1}
\definecolor{awesome}{rgb}{1.0, 0.13, 0.32}

\definecolor{figred}{rgb}{0.9, 0.1, 0.1}
\definecolor{figgreen}{rgb}{0.1, 0.7, 0.1}
\definecolor{figblue}{rgb}{0.1, 0.1, 0.9}
\definecolor{figmagenta}{rgb}{0.8, 0.1, 0.8}

\definecolor{myred}{rgb}{0.8, 0.0, 0.0}
\definecolor{myorange}{rgb}{0.8, 0.3, 0.0}
\definecolor{myyellow}{rgb}{0.5, 0.5, 0.0}
\definecolor{myyegr}{rgb}{0.2, 0.5, 0.0}
\definecolor{mygreen}{rgb}{0.0, 0.5, 0.0}
\definecolor{mygray}{rgb}{0.5, 0.5, 0.5}

\usepackage{relsize}
\usepackage{booktabs}
\usepackage{multirow}
\usepackage{xspace}
\usepackage{lipsum}

\usepackage{textcomp}
\newcommand{\mytexttilde}{\raisebox{0.5ex}{\texttildelow}}

\newcommand{\LBM}{LBM\xspace}

\newcommand{\method}{AnyView\xspace}
\newcommand{\ourbench}{AnyViewBench\xspace}

\newcommand{\tss}{\textsuperscript}
\newcommand{\tdag}{\tss{\textdagger}}
\newcommand{\tddag}{\tss{\textdaggerdbl}}
\newcommand{\tstar}{\tss{*}}

 \usepackage{colortbl}

\usepackage{eccvabbrv}

\usepackage{graphicx}
\usepackage{booktabs}

\usepackage[accsupp]{axessibility}  %

\usepackage[pagebackref,breaklinks,colorlinks,citecolor=eccvblue]{hyperref}

\usepackage{orcidlink}

\begin{document}

\title{AnyView:\\Synthesizing Any Novel View in Dynamic Scenes}

\titlerunning{AnyView: Synthesizing Any Novel View in Dynamic Scenes}

\author{
Basile Van Hoorick\inst{1} \and
Dian Chen\inst{1} \and
Shun Iwase\inst{1} \and
Pavel Tokmakov\inst{1} \and \\
Muhammad Zubair Irshad\inst{1} \and
Igor Vasiljevic\inst{1} \and
Swati Gupta\inst{1} \and
Fangzhou Cheng\inst{1,2} \and
Sergey Zakharov\inst{1} \and
Vitor Campagnolo Guizilini\inst{1}
}

\authorrunning{B.~Van Hoorick et al.}

\institute{Toyota Research Institute, CA, USA \and
Amazon Web Services, CA, USA\\[0.22cm]
{\larger \href{https://tri-ml.github.io/AnyView/}{tri-ml.github.io/AnyView}}}

\maketitle

\definecolor{Hgreen}{rgb}{0.0, 0.6, 0.0}
\definecolor{Hred}{rgb}{0.9, 0.1, 0.1}

\begin{center}
    \centering
    \vspace{-0.5mm}
    \includegraphics[width=\columnwidth]{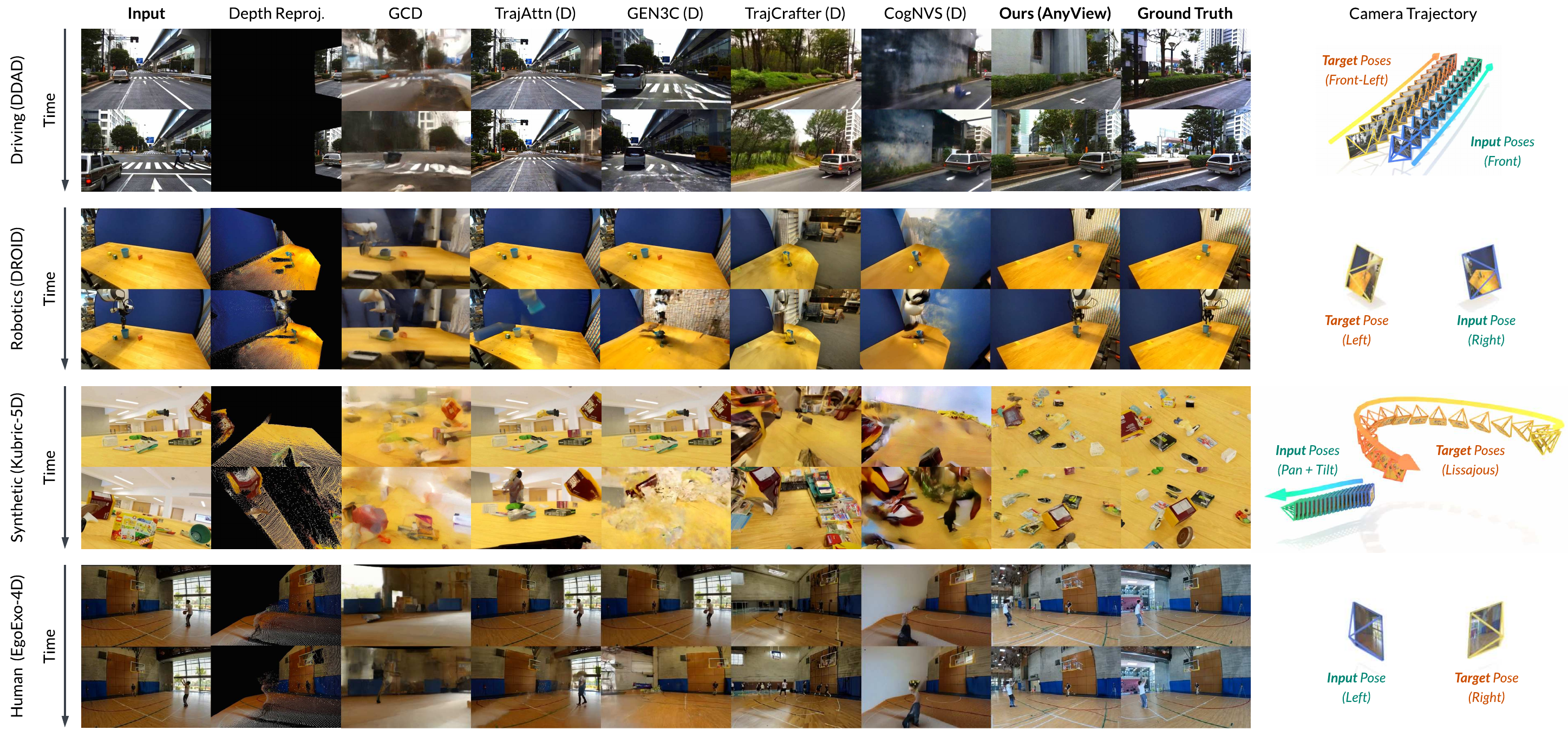}
    \captionsetup{width=\columnwidth}
    \vspace{-2mm}
    \captionof{figure}{
    \textbf{Enabling consistent extreme monocular dynamic view synthesis:}
    We introduce \emph{\method}, a diffusion framework that can generate videos of dynamic scenes from \emph{any} chosen perspective, 
    conditioned on a single input video.
    Our model operates end-to-end, without explicit scene reconstruction or expensive test-time optimization techniques.
    Existing methods tend to fail to extrapolate, largely copying the input view.
    More recent baselines can recover the overall structure in some cases (1st, 2nd rows), but fail when the camera trajectories become more complex (3rd row).
    Meanwhile, our method preserves scene geometry, appearance, and dynamics,
    despite working with drastically different target poses and highly ``incomplete'' visual observations.
    {\smaller{(D) indicates a baseline that relies on reprojected point clouds from estimated depth maps.}}
   \label{fig:teaser}
    \vspace{-1.5mm}
   }
\end{center}

\begin{abstract}
Modern generative video models excel at producing convincing, high-quality outputs, but struggle to maintain multi-view and spatiotemporal consistency in highly dynamic real-world environments.
In this work, we introduce \textbf{\method}, a diffusion-based video generation framework for \emph{dynamic view synthesis} with minimal inductive biases or geometric assumptions.
We leverage multiple data sources with various levels of supervision, including monocular (2D), multi-view static (3D) and multi-view dynamic (4D) datasets, to train a generalist spatiotemporal implicit representation capable of producing zero-shot novel videos from arbitrary camera locations and trajectories.
We evaluate \method on standard benchmarks, showing competitive results with the current state of the art, and propose \textbf{\ourbench}, a challenging new benchmark tailored towards \emph{extreme} dynamic view synthesis in diverse real-world scenarios.
In this more dramatic setting, we find that most baselines drastically degrade in performance, as they require significant overlap between viewpoints,
while \method maintains the ability to produce realistic, plausible, and spatiotemporally consistent videos when prompted from \emph{any} viewpoint.
\end{abstract}

\vspace{-4mm}
\section{Introduction}
\label{sec:intro}

Generating a new video from an arbitrary camera perspective while the scene is in motion is a highly ambitious and fundamentally under-constrained task.
A single input view only depicts a fraction of the world; the rest is occluded, transient, or simply unknown.
New moving objects may enter the scene at any moment, and unobserved regions might be dynamic themselves, further introducing uncertainty into the generative process.
Exact 4D reconstruction from such signals is therefore impractical in the general case.
For many downstream uses  of 4D video representations~\cite{liang2024dreamitate,wang2024drivedreamer, guo2025ctrlworldcontrollablegenerativeworld}, --- such as robotics, world models, simulation, telepresence, VR/AR, autonomous driving --- what matters is not an exact correspondence with ground truth, but rather whether the resulting representation is realistic, temporally stable, and self-consistent across large viewpoint changes.
A common problem with learned visuomotor policies, for example, is that they often suffer from brittleness under shifting camera poses~\cite{chen2024roviaug,tian2024view,pang2025learning,yang2025mobi}.

Humans routinely engage this problem in a way that is both rooted in intuition and very useful in practice: as we observe the physical world, we mentally ``re-project'' scenes, inferring likely layouts, object shapes, scene completions, and plausible dynamics from limited information~\cite{komatsu2006neural,nanay2018importance,park2007beyond,dicarlo2007untangling,baillargeon1987object,kahneman1992reviewing,shepard1971mental,burgess2006spatial}.
This is not simply a low-level reconstruction capability: it is essentially a powerful prior over shapes, semantics, materials, and motion that yields predictions that are largely viewpoint-invariant.
The goal of this paper is to take a step towards solving that objective: we target perceptually realistic 4D video synthesis under extreme camera trajectories and displacements.
To that end, we endow video generative models with the same inductive bias: to produce reasonable scene completions, based on a single input video,
that respect scene geometry, physics, and object permanence, even when there is little overlap with the conditioning view.

Most existing \emph{dynamic view synthesis} (DVS) approaches and benchmarks are not built for this regime~\cite{gao2022dynamic,som2024,gao2021dynamic,wu20244d,huang2025vivid4d}, as they typically operate in \emph{narrow} settings: the input and target cameras are spatially nearby, looking in similar directions, and thus methods are designed to maximize pixel metrics under limited motion, ignoring the rest of the scene. 
In particular, most current state of the art DVS methods~\cite{chen2025cognvs,mark2025trajectorycrafter,yesiltepe2025dynamicviewsynthesisinverse} rely on explicit 3D reconstructions (i.e., depth reprojection + image inpainting), costly test-time optimization and finetuning techniques, and support a limited set of camera trajectories.

To move away from this simplified setting, we first present \textbf{\method}, 
a novel diffusion-based DVS architecture for high-fidelity video-to-video synthesis under dramatic camera trajectory changes, capable of producing perceptually plausible and semantically consistent videos from arbitrary novel viewpoints.
Our framework is purposefully light on explicit inductive biases: camera parameters are provided via dense ray-space conditioning, allowing us to support any model (including non-pinhole), and the network learns to synthesize unobserved content implicitly, guided by large-scale, diverse training data. To reach this level of implicit 4D understanding, we leverage existing video foundation models as a source of rich internet-scale 2D appearance and motion priors, and augment them by incorporating multi-view geometry and camera controllability, learned using 12 multi-domain 3D and 4D datasets.

Secondly, due to the aforementioned shortcomings of existing evaluation procedures, we assemble \textbf{\ourbench}, 
a novel benchmark that formalizes and standardizes the \emph{extreme} DVS task across various domains (driving, robotics, and human activity), camera rigs (ego-centric and exo-centric), and camera motion patterns (fixed, linear, or complex, sometimes with changing intrinsics).
Each scene provides at least two time-synchronized views, enabling rigorous metric evaluations with ground truth videos without resorting to proxy setups.

Video results are best viewed on our project webpage: \href{https://tri-ml.github.io/AnyView/}{tri-ml.github.io/AnyView}.

\section{Related Work}
\label{sec:relwork}

\subsection{Video Generative Models}
In recent years, significant advances have been made in video generation, leading to the development of increasingly capable generative models.
Stability AI's SVD~\cite{blattmann2023stable} pioneered video diffusion by adding temporal layers to a pre-trained image diffusion network~\cite{rombach2022high}, allowing coherent short video clip generation from single images or text prompts.  
CogVideoX~\cite{yang2024cogvideox} introduced a 3D Variational Autoencoder (VAE) to compress videos across spatial and temporal dimensions, enhancing both compression rate and video fidelity. %
NVIDIA's Cosmos~\cite{agarwal2025cosmos} introduced a suite of models with strong long-range temporal consistency and flexible conditioning signals (text, image and video input).
Wan~\cite{wan2025wan} is a novel mixture of experts-based video generation architecture, and provides a suite of video world models that excel at prompt following and photorealistic generation.
However, none of these architectures were originally designed with camera conditioning in mind, focusing instead on future frame forecasting in the single -- or more recently multi-camera~\cite{nvidia2025worldsimulationvideofoundation} -- setting.

\subsection{Dynamic View Synthesis}

Dynamic view synthesis is the task of generating novel renderings from arbitrary viewpoints and timesteps given a monocular video of a dynamic scene.
A number of works have combined video generation with explicit geometric conditioning to improve geometric 3D consistency and control~\cite{zheng2024cami2v, he2024cameractrl, voleti2024sv3d, wu2025cat4d}.
Shape of Motion~\cite{som2024} addresses monocular dynamic reconstruction by representing scene motion through a compact set of SE(3) motion bases, enabling soft segmentation into multiple rigidly moving parts using monocular depth and long-range 2D tracks. It fuses monocular depth and long-range 2D tracks to obtain a globally consistent dynamic 3D representation.

\begin{figure*}[!t]
  \centering
  \includegraphics[width=1.00\linewidth]{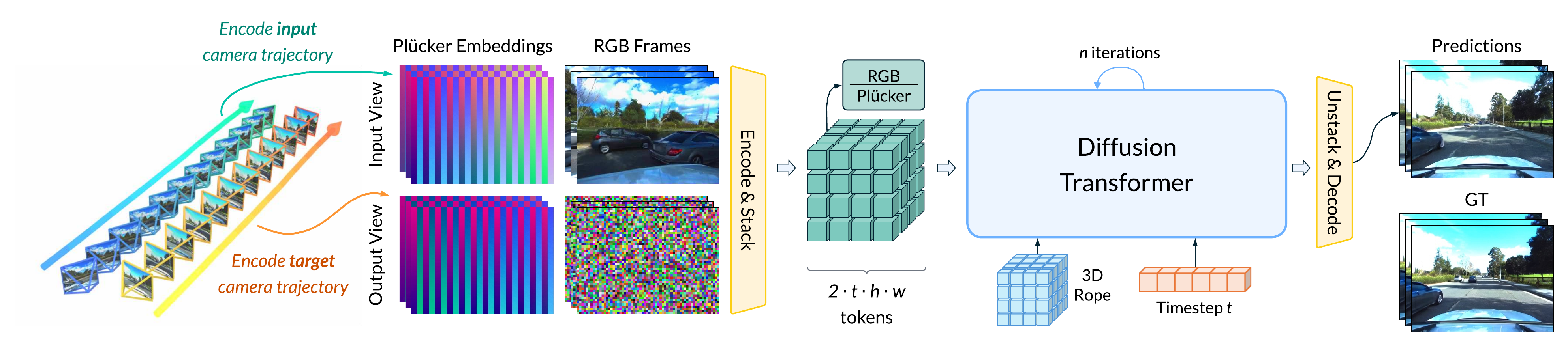}
  \vspace{-2mm}
  \caption{
  \textbf{The \method architecture.}
  For both the clean input and noisy target videos, we concatenate pixels (RGB values) and camera information (Pl\"{u}cker vectors) belonging to the \textit{same} viewpoint along the \textit{channel} dimension, after independently encoding each modality into latent embeddings.
  We then stack these two multimodal videos along the \textit{sequence dimension}, for a total of $2 \cdot t \cdot h \cdot w$ tokens, which are fed into the diffusion transformer to iteratively denoise the target video.
  }
  \label{fig:arch}
  \vspace{-2.5mm}
\end{figure*}

While explicit modeling approaches can achieve relatively high accuracy, they are computationally expensive and brittle. 
GCD~\cite{vanhoorick2024gcd} proposed to address dynamic view synthesis as an implicit problem, by re-purposing internet-scale video diffusion models via camera conditioning. This implicit formulation provides the greatest flexibility and robustness, but requires ground truth multi-view video data for training.
ReCamMaster~\cite{bai2025recammaster} advanced this research direction by utilizing a more powerful video generation model and a more realistic simulator to generate training data, whereas Trajectory Attention~\cite{xiao2025trajectory} augments video diffusion models with a trajectory-aware attention mechanism, improving fine-grained camera motion control and temporal consistency.
AC3D~\cite{bahmani2025ac3d} analyzes how video diffusion models internally represent 3D camera motion, adding ControlNet-style conditioning to improve controllability.  

Other methods~\cite{chen2025cognvs, mark2025trajectorycrafter,ren2025gen3c} have taken a hybrid approach by first lifting the input video in 3D via monocular depth estimation, reprojecting the resulting point cloud to the target camera pose, and then treating dynamic view synthesis as an inpainting problem. Among these methods, CogNVS~\cite{chen2025cognvs} further introduces test-time optimization to improve rendering accuracy at the cost of inference speed, while StreetCrafter~\cite{yan2024streetcrafter} focuses on autonomous driving scene generation, utilizing LiDAR renderings as the control signal. Very recently, InverseDVS~\cite{yesiltepe2025dynamic} has proposed a training-free approach that reformulates inpainting as structured latent manipulation in the noise initialization phase of a video diffusion model. 

While the shift towards explicit scene reconstruction and test-time optimization has led to high-quality dynamic view synthesis in the \textit{narrow} setting, where camera motion is limited to neighboring and highly overlapping regions, we experimentally demonstrate that these methods do not generalize to the more challenging \emph{extreme} setting. In contrast, data-driven, implicit approaches are in principle capable of dynamic view synthesis from any viewpoint, but are in practice limited by the availability of diverse training data. In this work, we address this by (1) combining a wide body of publicly available datasets to train \method — the first model capable of synthesizing arbitrary novel views in dynamic, real-world scenes; and (2) proposing a new benchmark, \ourbench, to properly evaluate dynamic view synthesis performance in this new setting.

\section{Methodology}
\label{sec:method}

\begin{figure*}[t!]
  \centering
  \includegraphics[width=1.00\textwidth]{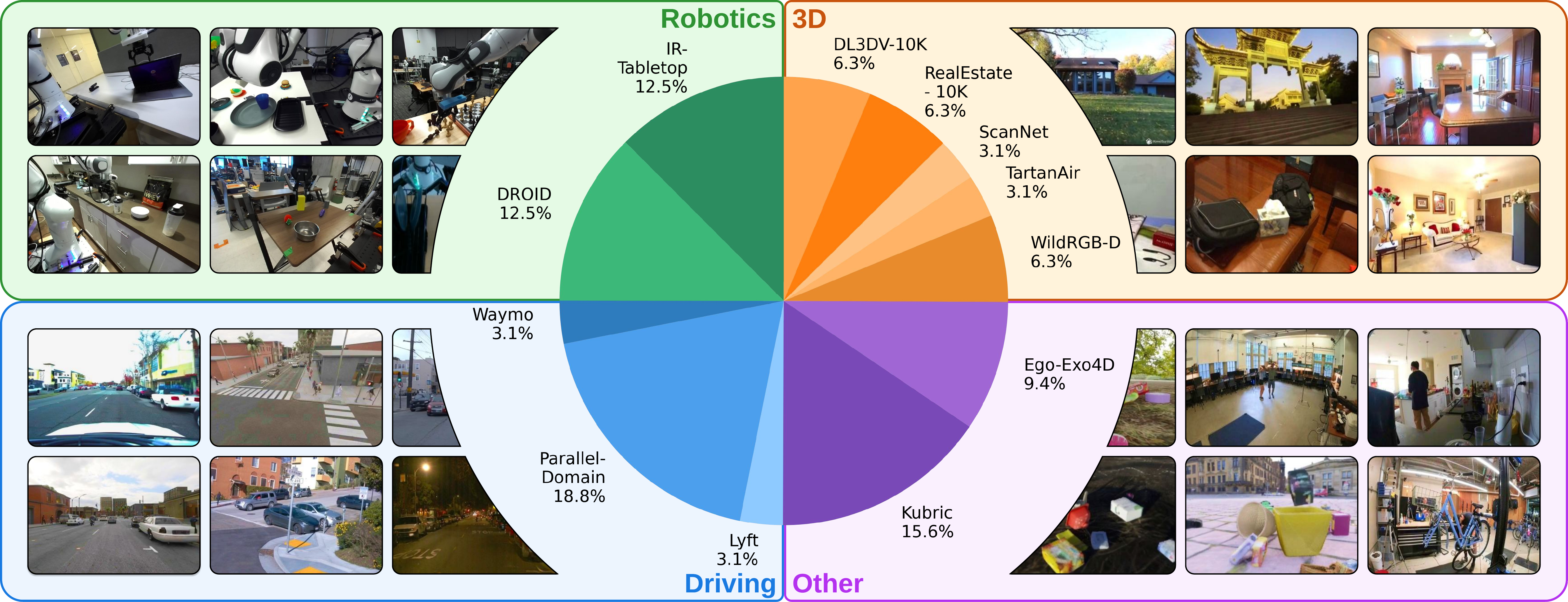}
  \caption{
  \textbf{Overview of our training data mixture.}
  We train and evaluate \method on both single-view and multi-view videos from four domains: \emph{3D}, \emph{Driving}, \emph{Robotics}, and \emph{Other} (see Section \ref{sec:datasets}).
  During training, we perform weighted sampling to ensure each domain is seen equally often, \ie comprises 25\% of the batch.
  }
  \label{fig:data}
\end{figure*}

\subsection{Problem Statement}

The goal of dynamic view synthesis (DVS) is to create an output video $\boldsymbol{V}_y$ of an underlying scene as depicted from a chosen virtual viewpoint $c_y$, given an input video $\boldsymbol{V}_x$ recorded by a camera with known poses $c_x$ and intrinsics $i_x$ over time.
Specifically, we define the input (observed) RGB video as $\boldsymbol{V}_{x} \in \mathbb{R}^{T \times H \times W \times 3}$, the target (unobserved) RGB video as $\boldsymbol{V}_{y} \in \mathbb{R}^{T \times H \times W \times 3}$, the input camera trajectory as $c_{x} \in \mathbb{R}^{T \times 4 \times 4}$ with intrinsics $i_x \in \mathbb{R}^{T \times 3 \times 3}$, and the target camera trajectory as $c_{y} \in \mathbb{R}^{T \times 4 \times 4}$ with intrinsics $i_y \in \mathbb{R}^{T \times 3 \times 3}$.
Using a generative model $f$, we estimate $\boldsymbol{V}_y$ corresponding to the desired novel viewpoint $c_y$ by drawing from a conditional probability distribution:
\begin{align}
    \boldsymbol{V}_y
    \sim P_f \left( \boldsymbol{V}_y \mid \boldsymbol{V}_x,c_x,i_x,c_y,i_y \right)
\end{align}
The camera parameters $c_x,i_x,c_y,i_y$ represent two sequences of fully specified 6-DoF $SE(3)$ camera poses, ensuring that the task setting is both general and unambiguous.
Moreover, there should be some spatial overlap in content between the two perspectives $c_x$ and $c_y$ (even if this overlap is temporally asynchronous), otherwise the conditioning signal loses its relevance.

\subsection{Architecture}

The task described above involves (1) \emph{synthesis of high-dimensional data} in the form of multiple images, and (2) \emph{considerable uncertainty handling} mainly because of occlusion and ambiguous object motion. 
These requirements are challenging, but naturally lend themselves to being implemented using the generative video paradigm.
Hence, we adopted Cosmos~\cite{nvidia2025worldsimulationvideofoundation}, a latent diffusion transformer, as our underlying base representation, for its efficiency, high-quality pretrained checkpoints, and flexible conditioning mechanisms (\eg text, image, and video).

Our proposed \method architecture, illustrated in Figure~\ref{fig:arch}, prioritizes simplicity and scalability.
Contrary to most state-of-the-art methods~\cite{ren2025gen3c,chen2025cognvs,xiao2025trajectory,mark2025trajectorycrafter}, we elect to not use warped depth maps as explicit conditioning, owing to the risk of compounding errors from depth estimation, and instead rely solely on a learned implicit representation as our rendering mechanism.
The reasoning behind this decision is so that we can achieve \textit{unbounded} dynamic view synthesis, that does not require substantial overlap between target and generated videos, thus allowing for more extreme camera motion.
We explore this property in our proposed \ourbench, outperforming baselines that rely on explicit reprojection. %

In order to make \method 4D-aware and controllable, we feed information about both viewpoints into the network in a structured yet straightforward way. To account for the possible lack of an absolute frame of reference, all camera poses are expected to exist relative to the target viewpoint $c_{y,0}$ at time $t=0$.
In other words, $c_y$ always starts at the ``origin", with $c_{y,0}=I_{4 \times 4}$,
mapping to the identity matrix.
If this is not the case, a simple change of coordinate system can be done by applying $\tilde{c} = c \cdot c_{y,0}^{-1}$, assuming the camera-to-world extrinsics convention.

First, the given video $\boldsymbol{V}_x$ is compressed into a latent space by a video tokenizer to become $\boldsymbol{v}_x \in \mathbb{R}^{t \times h \times w \times d}$, with spatiotemporal downsampling ratios $T/t=4$ and $H/h=W/w=8$, and embedding size $d=16$. %
We then encode all camera parameters $c_x,i_x,c_y,i_y$ into a unified \emph{Pl\"{u}cker representation} $\boldsymbol{P}=(\boldsymbol{r},\boldsymbol{m})$~\cite{Hodge1947}, which combines extrinsics and intrinsics into a dense map containing per-pixel \emph{ray} vectors $\boldsymbol{r}$ and \emph{moment} vectors $\boldsymbol{m}=\boldsymbol{r} \times \boldsymbol{o}$.
This results in two quantities $\boldsymbol{P}_x,\boldsymbol{P}_y \in \mathbb{R}^{T \times H \times W \times 6}$, \ie tensors with the same dimensionality as a 6-channel video, or two 3-channel videos. We can therefore separately tokenize the rays $\boldsymbol{r} \in \mathbb{R}^{T \times H \times W \times 3}$ and moments $\boldsymbol{m} \in \mathbb{R}^{T \times H \times W \times 3}$ (shown as alternating columns in Figure~\ref{fig:arch}) the same way as before into $\boldsymbol{p}_x,\boldsymbol{p}_y \in \mathbb{R}^{t \times h \times w \times 2d}$. An interesting property of using Pl\"{u}cker maps instead of direct camera conditioning~\cite{vanhoorick2024gcd} is the natural handling of non-pinhole camera models, since the 3D ray vectors directly capture camera intrinsics in a general, non-parametric way.

Because latent RGB and Pl\"{u}cker tokens from each viewpoint contain information pertaining to the same spatiotemporal region, we merge them via concatenation along the channel dimension, while keeping tokens from separate viewpoints separate.
Since there are two viewpoints in total, this results in a sequence of $2 \cdot t \cdot h \cdot w$ tokens, each of length $3 \cdot d$.
All tokens are tagged with rotary positional embeddings~\cite{su2021roformer}, as well as a unique per-view embedding.
After completing all self-attention and cross-attention blocks, the output sequence is the latent RGB video $\boldsymbol{v}_y \in \mathbb{R}^{t \times h \times w \times d}$.  During training, these latent tokens are supervised with an $\mathcal{L}_2$ loss, and during inference, they are iteratively denoised before finally being decoded into a generated video $\boldsymbol{V}_y \in \mathbb{R}^{T \times H \times W \times 3}$.

\begin{table*}[t!]
  \centering
  \smaller
  \setlength{\tabcolsep}{0.3pt}
  \renewcommand{\arraystretch}{0.99}
  
\scalebox{0.84}{%
  \begin{tabular}{@{}lcccccccc@{}}
    \toprule
    Benchmark
    & S/R & Domain
    & Type
    & \#Cameras & \#Episodes & Resolution
    & Input Cam.
    & Gen. Type
    \\

    \midrule
    \midrule
    \multicolumn{9}{@{}l }
    {\textbf{Narrow} dynamic view synthesis} \\
    \midrule
    \midrule
    
    DyCheck iPhone~\cite{gao2022dynamic}
    & Real & Hand-Object
    & 4D
    & 2 -- 3
    & 5 / 7
    & $288 \times 384$
    & Moving
    & 0-shot overall
    \\
    
    Kubric-4D (gradual)~\cite{greff2021kubric}
    & Sim & Multi-Object
    & 4D
    & 16 %
    & 20 / 100
    & $384 \times 256$
    & Fixed
    & In-dist.
    \\
    
    ParDom-4D (gradual)~\cite{parallel_domain}
    & Sim & Driving
    & 4D
    & 19 %
    & 20 / 61
    & $384 \times 256$
    & Variable
    & In-dist.
    \\

    \midrule
    \midrule
    \multicolumn{9}{@{}l }
    {\textbf{\textit{\ourbench}: In-distribution extreme} dynamic view synthesis} \\
    \midrule
    \midrule

    DROID (ID)~\cite{droid}
    & Real & Robotics
    & 4D
    & 2 %
    & 64 / 3,301
    & $384 \times 208$
    & Fixed
    & In-dist.
    \\
    
    Ego-Exo4D (ID)~\cite{egoexo4d}
    & Real & Human Activity
    & 4D
    & 4 -- 5 %
    & 64 / 276
    & $384 \times 208$
    & Fixed
    & In-dist.
    \\
    
    \LBM
    & Sim + Real & Robotics
    & 4D
    & 2 %
    & 64 / 5,988
    & $336 \times 256$
    & Fixed
    & In-dist.
    \\
    
    Kubric-4D (direct)~\cite{greff2021kubric}
    & Sim & Multi-Object
    & 4D
    & 16 %
    & 20 / 100
    & $384 \times 256$
    & Fixed
    & In-dist.
    \\
    
    Kubric-5D~\cite{greff2021kubric}
    & Sim & Multi-Object
    & 4D
    & 16 %
    & 64 / 200
    & $384 \times 256$
    & Variable
    & In-dist.
    \\
    
    Lyft~\cite{lyftl5}
    & Real & Driving
    & 4D
    & 6 %
    & 64 / 436
    & $384 \times 320$
    & Variable
    & In-dist.
    \\
    
    ParDom-4D (direct)~\cite{parallel_domain}
    & Sim & Driving
    & 4D
    & 19 %
    & 20 / 61
    & $384 \times 256$
    & Variable
    & In-dist.
    \\
    
    Waymo~\cite{waymo}
    & Real & Driving
    & 4D
    & 5 %
    & 64 / 202
    & $384 \times \{176,256\}$
    & Variable
    & In-dist.
    \\

    \midrule
    \midrule
    \multicolumn{9}{@{}l }
    {\textbf{\textit{\ourbench}: Zero-shot extreme} dynamic view synthesis} \\
    \midrule
    \midrule

    Argoverse~\cite{Argoverse2}
    & Real & Driving
    & 4D
    & 7 %
    & 64 / 1,042
    & $\{ 288,384 \} \times \{ 288,384 \}$
    & Variable
    & dataset
    \\

    AssemblyHands~\cite{ohkawa:cvpr23}
    & Real
    & Hand-Object
    & 4D
    & 8 %
    & 20 / 20
    & $384 \times 208$
    & Fixed
    & domain
    \\

    DDAD~\cite{ddad}
    & Real & Driving
    & 4D
    & 6 %
    & 64 / 200
    & $384 \times 240$
    & Variable
    & dataset
    \\

    DROID (OOD)~\cite{droid}
    & Real & Robotics
    & 4D
    & 2 %
    & 64 / 252
    & $384 \times 208$
    & Fixed
    & station
    \\
    
    Ego-Exo4D (OOD)~\cite{egoexo4d}
    & Real & Human Activity
    & 4D
    & 4 -- 5 %
    & 64 / 408
    & $384 \times 208$
    & Fixed
    & activity/site
    \\

    \bottomrule
  \end{tabular}
}

  \vspace{2mm}
  \caption{
  \textbf{Testing datasets.}
  We evaluate on several benchmarks that cover both \emph{narrow} and \emph{extreme} settings.
  We define \textbf{\ourbench} as a multi-faceted benchmark focusing on the latter category, setting a new standard for consistent dynamic view synthesis in challenging settings.
Test splits are capped at 64 per dataset with uniform subsampling. 
\emph{Input Cam.} refers to what the camera characteristic of the observed video (\ie static vs dynamic).
  The rightmost column (\emph{Generalization Type}) qualitatively denotes how large the distribution shift is relative to the \method training mixture.
  }
  \label{tab:data_eval}
  \vspace{-6mm}
  
\end{table*}

\subsection{Datasets}
\label{sec:datasets}

Because \method does not rely on any explicit conditioning mechanism (\eg intermediate depth maps) to facilitate the rendering of novel viewpoints, it must learn implicit multi-view geometry as well as a wide range of appearance priors, to be able to inpaint and outpaint potentially large unobserved portions of the scene.
In order to train such a generalist spatiotemporal representation capable of handling multiple domains, we combined 12 different 4D datasets into our unified training pipeline.
Among them is \emph{Kubric-5D}, our newly introduced variation of Kubric-4D~\cite{vanhoorick2024gcd,greff2021kubric} that vastly increases the diversity of camera trajectories. %
We classify our training datasets into four distinct quadrants: \textit{Robotics}, \textit{Driving}, \textit{3D}, and \textit{Other}.
A visual overview is illustrated in Figure~\ref{fig:data}, and more details are provided in the supplementary material. %
To the best of our knowledge, this data mixture covers a significant portion of publicly available multi-view video datasets.
We leave the inclusion of additional 4D data~\cite{zheng2023point,infinigen2023infinite,infinigen2024indoors} to future work.

\subsection{Implementation Details}

We train \method for 40,000 iterations on 64 NVIDIA H200 GPUs at a global batch size of 512.
We apply curriculum learning with increasing resolution:
first we train at a largest image dimension of $384$ for 30,000 steps, before finetuning at a largest image dimension of $576$.
The initial learning rate is $5 \cdot 10^{-5}$, and drops smoothly to $1 \cdot 10^{-5}$ according to a cosine schedule.
All experiments are performed with the \texttt{Cosmos-Predict2-2B-Video2World}~\cite{cosmos-predict2-2025} model,
starting from their pretrained network, which has around 2 billion parameters.
We disable language conditioning, since it is not relevant to our task setting.
Furthermore, in order to properly combine datasets with varying physical scales, we divide the translation vectors of all cameras $\{c_x,c_y\}$ by a carefully chosen per-dataset normalization constant to ensure the resulting Pl\"{u}cker values always fall in the range $[-1, 1]$, occasionally clipping pixels as needed.

\section{Experiments}
\label{sec:exp}

\subsection{Evaluation Challenges}

As the field is evolving, many existing DVS benchmarks are beginning to \textbf{lack difficulty},
containing scenes with minimal object motion and modest camera transformations~\cite{gao2022dynamic,yoon2020novel,lin2021deep,vanhoorick2024gcd}.
Qualitative results are often demonstrated on camera trajectories with rotational variations of only about $10-30$ degrees relative to the center of the scene~\cite{som2024,vanhoorick2024gcd,zhang2024recapture,bai2025recammaster,yesiltepe2025dynamicviewsynthesisinverse}.
Consequently, the heavy lifting of inpainting large occlusions is mostly avoided, making it unclear to which extent these models learn robust, multi-view consistent 4D representations.
These efforts are further complicated by a \textbf{lack of standardization}, which 
can be partially attributed to the inherent complexity of DVS: merely describing the task is insufficient to define a path towards practical execution.
Design choices often left in the dark include but are not limited to: video resolution, number of frames, camera controllability and conventions, used frames of reference, the space of possible camera transformations, and so on.

\subsection{Benchmarks}

\begin{table}[t!]
  \centering
  \smaller
  \setlength{\tabcolsep}{1pt}
  \renewcommand{\arraystretch}{0.99}

  \begin{tabular}{@{}l|ccc|c|c|c|c}
    \toprule
    Method
    & PSNR$\uparrow$ & SSIM$\uparrow$ & LPIPS$\downarrow$
    & TTO
    & Aux.
    & 0-shot
    & Time  %
    \\

    \midrule
    \midrule
    \multicolumn{5}{@{}l }
    {\textbf{DyCheck iPhone}~\cite{gao2022dynamic}} 
    \\
    \midrule
    \midrule

    {ShapeOfMotion\tdag~\cite{som2024}}
    & 16.72
    & 0.630
    & 0.450
    & \checkmark
    & D P T
    & --
    & \color{myred}\mytexttilde 1h
    \\

    CogNVS\tdag~\cite{chen2025cognvs}
    & 16.94
    & 0.449
    & 0.598
    & \checkmark
    & D(GT)
    & --
    & \color{myred}\mytexttilde 1h
    \\

    \midrule

    GEN3C\tstar~\cite{ren2025gen3c}
    & 10.13 & 0.175 & 0.695
    & & D
    & \checkmark
    & \color{myorange} \mytexttilde 15m
    \\

    TrajAttn\tstar~\cite{xiao2025trajectory}
    & 10.30 & 0.181 & 0.682
    & & D
    & \checkmark
    & \color{myorange} \mytexttilde 10m
    \\

    \midrule
    TrajCrafter\tdag~\cite{mark2025trajectorycrafter}
    & 14.24 & 0.417 & 0.519  %
    & & D
    & \checkmark
    & \color{mygreen} \mytexttilde 10s
    \\

    \midrule

    GCD\tstar~\cite{vanhoorick2024gcd}
    & 11.43 & 0.247 & 0.728 %
    & & --
    & \checkmark
    & \color{mygreen} \mytexttilde 10s
    \\

    \textit{\method}
    & 13.47 & 0.295 & 0.550
    & & --
    & \checkmark
    & \color{mygreen} \mytexttilde 10s
    \\

    \rowcolor{gray!10}
    \textit{\quad w/ GCD data}
    & 11.38 & 0.191 & 0.662
    & & --
    & \checkmark
    & \color{mygreen} \mytexttilde 10s
    \\

    \rowcolor{gray!10}
    \textit{\quad from scratch}
    & 9.40 & 0.173 & 0.732
    & & --
    & \checkmark
    & \color{mygreen} \mytexttilde 10s
    \\

    \midrule
    \midrule
    \multicolumn{5}{@{}l }
    {\textbf{Kubric-4D} (gradual)~\cite{greff2021kubric}} \\
    \midrule
    \midrule

    CogNVS\tdag~\cite{chen2025cognvs}
    & 22.63
    & 0.760
    & 0.232
    & \checkmark & D(GT)
    & --
    & \color{myred} \mytexttilde 1h
    \\

    ReCapture\tdag~\cite{zhang2024recapture}
    & 20.92
    & 0.596
    & 0.402
    & \checkmark & D
    & --
    & \color{myorange} \mytexttilde 15m
    \\

    \midrule

    GEN3C\tdag~\cite{ren2025gen3c}
    & 19.41 & 0.630 & 0.290
    & & D
    & \checkmark
    & \color{myorange} \mytexttilde 15m
    \\

    TrajAttn\tstar~\cite{xiao2025trajectory}
    & 15.73 & 0.404 & 0.530
    & & D
    & \checkmark
    & \color{myyellow} \mytexttilde 5m
    \\

    \midrule

    TrajCrafter\tddag~\cite{chen2025cognvs}
    & 20.93 & 0.730 & 0.257
    & & D
    & \checkmark
    & \color{mygreen} \mytexttilde 10s
    \\

    \midrule

    GCD\tstar~\cite{vanhoorick2024gcd}
    & 20.42 & 0.581 & 0.405  %
    & & --
    &
    & \color{mygreen} \mytexttilde 10s
    \\

    \textit{\method}
    & 21.21 & 0.644 & 0.358
    & & --
    &
    & \color{mygreen} \mytexttilde 10s
    \\

    \rowcolor{gray!10}
    \textit{\quad w/ GCD data}
    & 22.06 & 0.611 & 0.343
    & & --
    &
    & \color{mygreen} \mytexttilde 10s
    \\

    \rowcolor{gray!10}
    \textit{\quad from scratch}
    & 17.68 & 0.445 & 0.499
    & & --
    &
    & \color{mygreen} \mytexttilde 10s
    \\

    \midrule
    \midrule
    \multicolumn{5}{@{}l }
    {\textbf{ParDom-4D} (gradual)~\cite{parallel_domain}} \\
    \midrule
    \midrule

    CogNVS\tdag~\cite{chen2025cognvs}
    & 24.34
    & 0.797
    & 0.302
    & \checkmark
    & D(GT)
    & --
    & \color{myred} \mytexttilde 1h
    \\

    \midrule

    GEN3C\tstar~\cite{ren2025gen3c}
    & 18.40 & 0.528 & 0.542
    & & D
    & \checkmark
    & \color{myorange} \mytexttilde 15m
    \\

    TrajAttn\tstar~\cite{xiao2025trajectory}
    & 20.03 & 0.566 & 0.518
    & & D
    & \checkmark
    & \color{myyellow} \mytexttilde 5m
    \\

    \midrule

    TrajCrafter\tddag~\cite{chen2025cognvs}
    & 21.46 & 0.719 & 0.342
    & & D
    & \checkmark
    & \color{mygreen} \mytexttilde 10s
    \\

    \midrule

    GCD\tstar~\cite{vanhoorick2024gcd}
    & 24.75 & 0.724 & 0.355  %
    & & --
    &
    & \color{mygreen} \mytexttilde 10s
    \\

    \textit{\method}
    & 26.29 & 0.758 & 0.320
    & & --
    &
    & \color{mygreen} \mytexttilde 10s
    \\

    \rowcolor{gray!10}
    \textit{\quad w/ GCD data}
    & 26.18 & 0.748 & 0.345
    & & --
    &
    & \color{mygreen} \mytexttilde 10s
    \\

    \rowcolor{gray!10}
    \textit{\quad from scratch}
    & 21.89 & 0.590 & 0.507
    & & --
    &
    & \color{mygreen} \mytexttilde 10s
    \\

    \bottomrule
  \end{tabular}

  \vspace{2mm}
  \caption{
  \textbf{Narrow DVS results.}
  We compare against several state-of-the-art baselines, including those using test-time optimization (TTO) and auxiliary networks (Aux.) for depth (D), poses (P), and/or 2D point tracks (T).
  {\smaller
 The inference runtime assumes that a video was not observed before, and thus includes a test-time optimization stage if present.
 Results reported by:
  \tdag original paper;
  \tddag another paper (cited);
  \tstar computed by us.
  }
  \colorbox{gray!10}{Gray} rows are ablations of \method.
  \vspace{-2mm}
  }
  \label{tab:res_nar_psl}

\end{table}

\begin{figure*}[t!]
  \centering
  \includegraphics[width=0.49\linewidth]{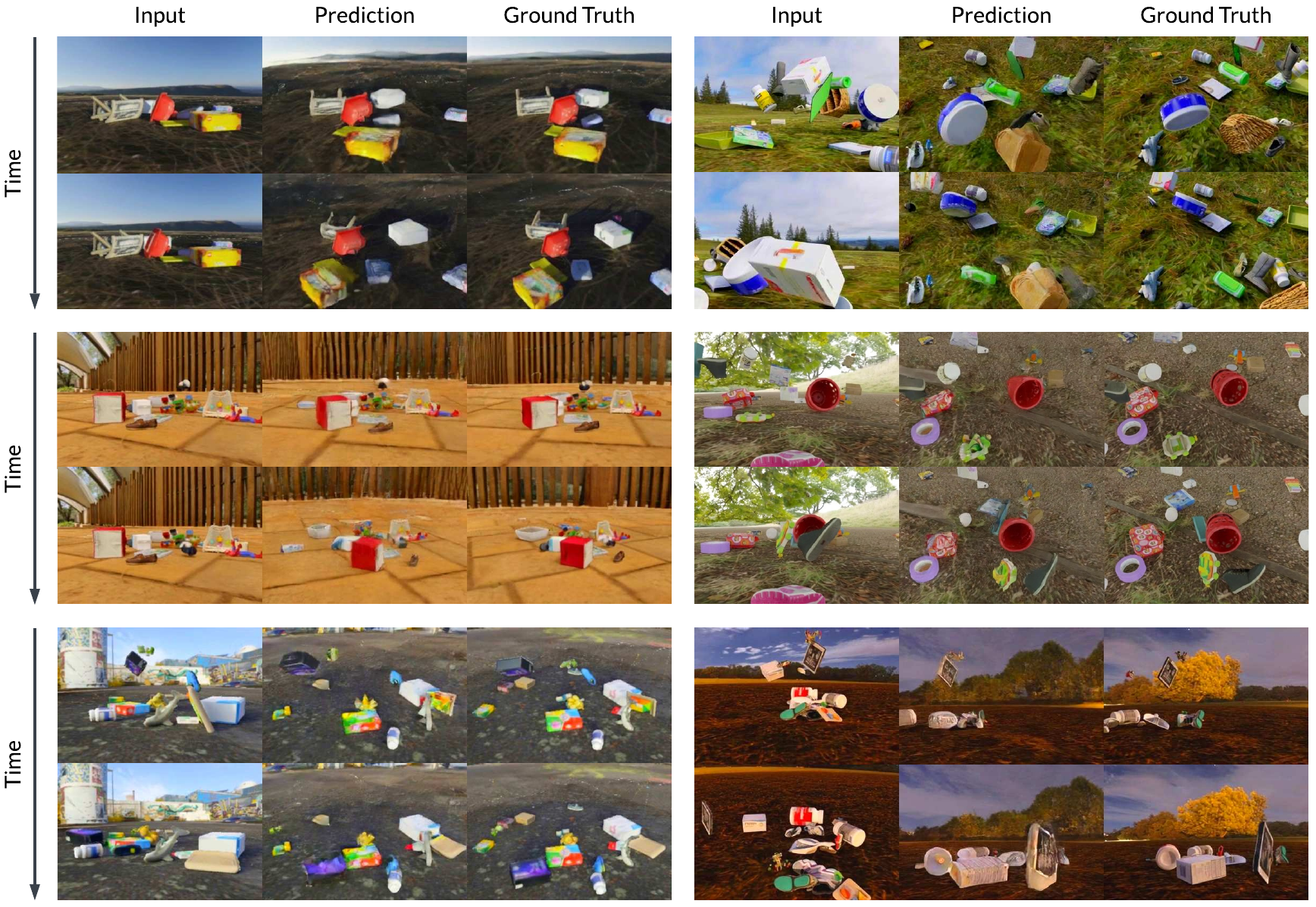}
  \includegraphics[width=0.49\linewidth]{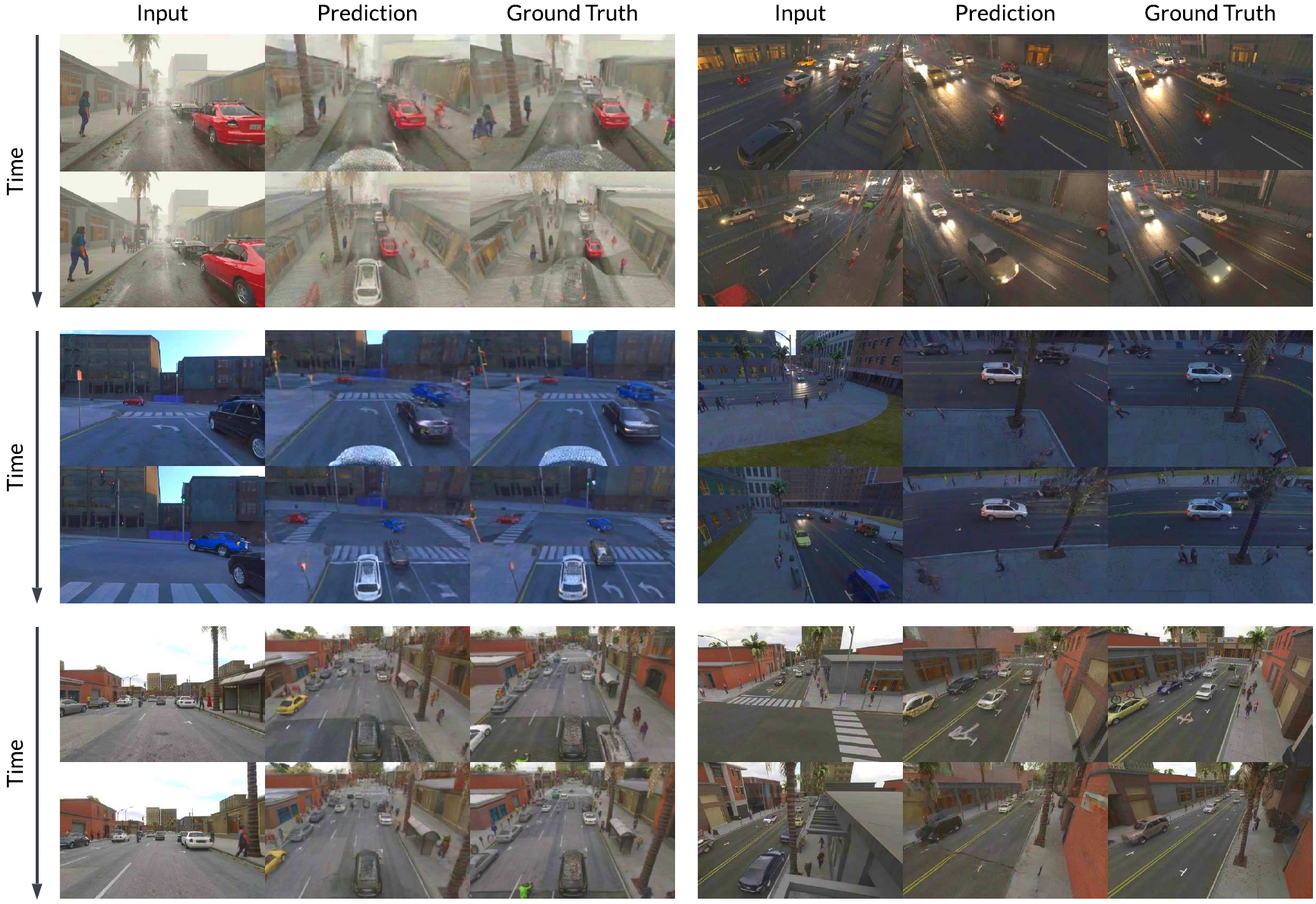}
  \vspace{-0.5mm}
  \caption{
  \textbf{\method in-domain DVS results}
  on \textbf{Kubric-4D} (left) and \textbf{Pardom-4D} (right).
  We show the first and last frame of each video.
  The scene layout is generally preserved very well, despite drastic viewpoint changes and/or heavy occlusion from the input vantage point.
  }
  \label{fig:res_pardom}
\vspace{-2.5mm}
\end{figure*}

\begin{figure}
  \centering
  \includegraphics[width=0.8\linewidth]{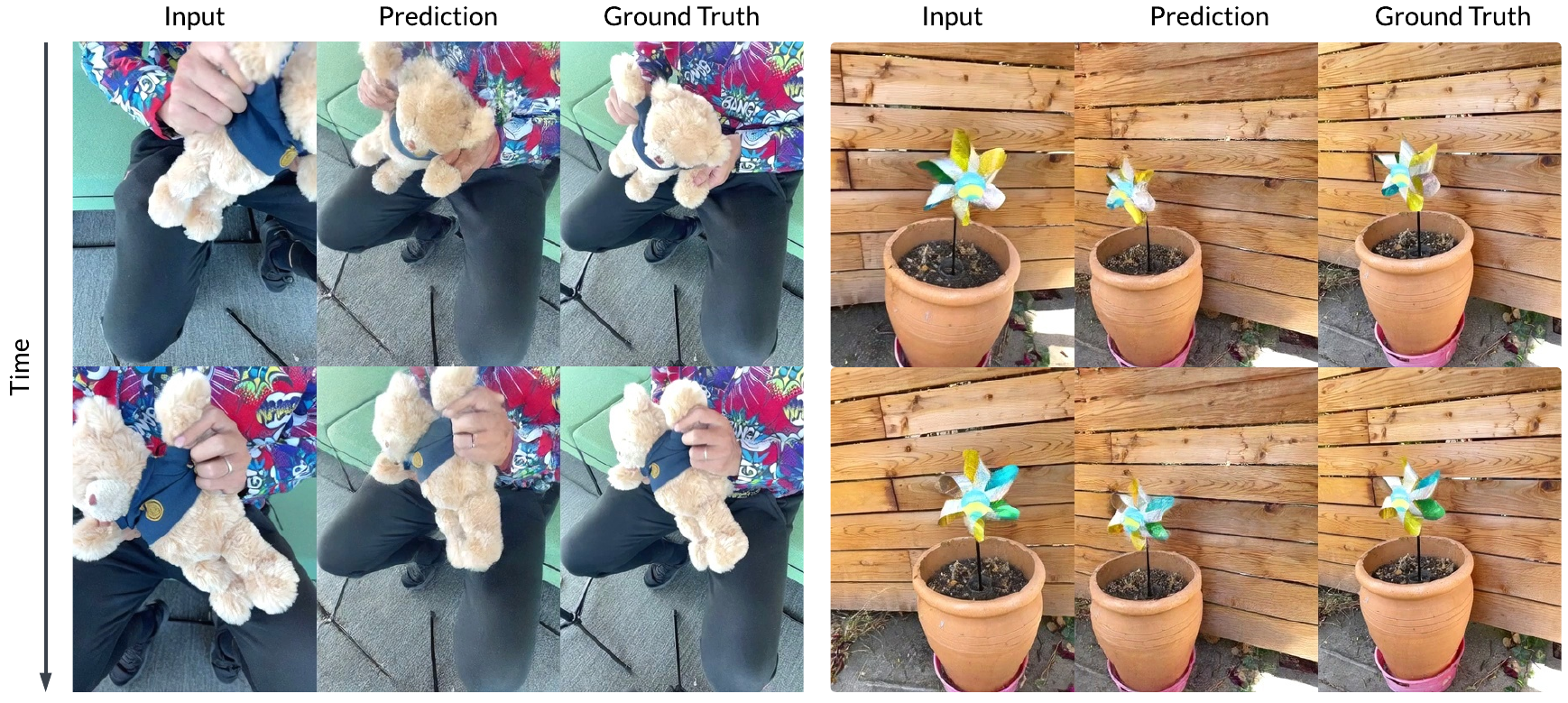}
  \vspace{-0.5mm}
  \caption{
  \textbf{Results on DyCheck iPhone (0-shot narrow DVS).}
  While not highly dynamic, these scenes contain subtle, intricate motions and hand-object interactions. %
  }
  \label{fig:res_dycheck}
\end{figure}

We first consider three popular DVS benchmarks that can be classified as falling into the ``narrow'' regime.
Then, to address the aforementioned concerns, we propose \emph{\ourbench}, which substantially pushes models into the more challenging ``extreme'' regime. %

\noindent\textbf{DyCheck iPhone (narrow DVS).}
The iPhone dataset~\cite{gao2022dynamic} is a small collection of high-quality, real-world, multi-view videos of easy-to-moderate difficulty established to measure DVS fidelity.
Following previous works~\cite{chen2025cognvs}, that have pointed out that the provided camera poses are not very accurate, we compute corrected extrinsics using MoSca~\cite{lei2025mosca}.

\noindent\textbf{Kubric-4D and ParDom-4D (narrow + extreme DVS).}
The GCD~\cite{vanhoorick2024gcd} paper introduced two synthetic datasets for DVS training and evaluation, based on the Kubric~\cite{greff2021kubric} and ParallelDomain~\cite{parallel_domain} simulation environments.

\begin{table*}
  \centering
  \smaller
  \setlength{\tabcolsep}{1pt}
  \renewcommand{\arraystretch}{0.99}

\scalebox{0.97}{%
  \begin{tabular}{@{}l |cccc|cccc|cccc@{}} %
    \toprule

    Dataset
    & PSNR$\uparrow$ & SSIM$\uparrow$ & LPIPS$\downarrow$ & FVD$\downarrow$
    & PSNR$\uparrow$ & SSIM$\uparrow$ & LPIPS$\downarrow$ & FVD$\downarrow$
    & PSNR$\uparrow$ & SSIM$\uparrow$ & LPIPS$\downarrow$ & FVD$\downarrow$
    \\

    \midrule
    \midrule
    \multicolumn{12}{@{}l }
    {\textbf{\textit{\ourbench}: In-distribution}} \\
    \midrule
    \midrule

    & \multicolumn{4}{c|}{GCD~\cite{vanhoorick2024gcd}}
    & \multicolumn{4}{c|}{TrajAttn~\cite{xiao2025trajectory}}
    & \multicolumn{4}{c}{GEN3C~\cite{ren2025gen3c}}
    \\

    \cmidrule(lr){2-5}\cmidrule(lr){6-9}\cmidrule(lr){10-13}

    DROID~\cite{droid}
    & 10.18 & 0.255 & 0.688 & 5699
    & 9.93 & 0.247 & 0.671 & 3430
    & 9.62 & 0.211 & 0.666 & \underline{3359}
    \\

    EgoExo4D~\cite{egoexo4d}
    & \underline{12.10} & \underline{0.255} & 0.670 & 3353
    & 11.64 & 0.234 & 0.646 & 2520
    & 11.69 & 0.222 & 0.643 & \underline{2097}
    \\

    \LBM
    & 12.59 & 0.398 & 0.694 & 2458
    & 13.47 & 0.421 & 0.614 & \underline{1891}
    & 13.48 & 0.449 & 0.581 & 1943
    \\

    Kubric-4D~\cite{greff2021kubric}
    & \underline{17.57} & \underline{0.477} & \underline{0.512} & \underline{3731}
    & 13.47 & 0.320 & 0.607 & 5257
    & 13.59 & 0.341 & 0.599 & 4556
    \\

    Kubric-5D
    & \underline{13.83} & \underline{0.391} & 0.644 & 3405
    & 13.25 & 0.360 & 0.628 & 3661
    & 13.10 & 0.327 & 0.627 & \underline{3241}
    \\

    Lyft~\cite{lyftl5}
    & 8.72 & \underline{0.335} & 0.697 & 6442
    & 8.33 & 0.273 & \underline{0.621} & \underline{1819}
    & 8.43 & 0.286 & 0.634 & 2211
    \\

    ParDom-4D~\cite{parallel_domain}
    & \underline{22.67} & \textbf{0.656} & \textbf{0.457} & \underline{3579}
    & 16.91 & 0.445 & 0.610 & 6027
    & 16.64 & 0.478 & 0.590 & 5738
    \\

    Waymo~\cite{waymo}
    & 12.93 & \underline{0.393} & 0.647 & 4344
    & 12.55 & 0.350 & 0.613 & \underline{1617}
    & 12.98 & 0.350 & 0.600 & 1653
    \\

    \midrule

    Average
    & \underline{13.95} & \underline{0.400} & 0.623 & 4126
    & 12.44 & 0.331 & 0.626 & 3278
    & 12.44 & 0.333 & 0.617 & \underline{3100}
    \\
    
    \midrule
   & \multicolumn{4}{c|}{TrajCrafter~\cite{mark2025trajectorycrafter}}
   & \multicolumn{4}{c|}{CogNVS~\cite{chen2025cognvs}}
   & \multicolumn{4}{c}{Ours (\method)}
    \\
    \cmidrule(lr){2-5}\cmidrule(lr){6-9}\cmidrule(lr){10-13}

    DROID~\cite{droid}
    & \underline{10.45} & 0.257 & \underline{0.620} & 3416
    & 9.44 & \underline{0.281} & 0.634 & 4468
    & \textbf{14.42} & \textbf{0.443} & \textbf{0.473} & \textbf{1399}  %
    \\

    EgoExo4D~\cite{egoexo4d}
    & 11.33 & 0.195 & \underline{0.642} & 3367
    & 10.80 & 0.241 & 0.670 & 2607
    & \textbf{17.46} & \textbf{0.493} & \textbf{0.410} & \textbf{995}  %
    \\

    \LBM
    & \underline{13.68} & 0.447 & \underline{0.537} & 2090
    & 13.30 & \underline{0.453} & 0.548 & 1903
    & \textbf{17.63} & \textbf{0.619} & \textbf{0.386} & \textbf{594}  %
    \\

    Kubric-4D~\cite{greff2021kubric}
    & 14.14 & 0.294 & 0.592 & 5357
    & 12.55 & 0.329 & 0.601 & 5104
    & \textbf{19.91} & \textbf{0.541} & \textbf{0.427} & \textbf{2009}  %
    \\

    Kubric-5D
    & 13.30 & 0.287 & \underline{0.625} & 3450
    & 12.18 & 0.318 & 0.643 & 4563
    & \textbf{17.04} & \textbf{0.460} & \textbf{0.488} & \textbf{1130}  %
    \\

    Lyft~\cite{lyftl5}
    & \underline{8.73} & 0.251 & \underline{0.621} & 2444
    & 8.34 & 0.319 & 0.628 & 3462
    & \textbf{15.29} & \textbf{0.560} & \textbf{0.389} & \textbf{719}  %
    \\

    ParDom-4D~\cite{parallel_domain}
    & 18.23 & 0.475 & 0.586 & 5790
    & 18.36 & 0.499 & 0.564 & 5668
    & \textbf{22.85} & \underline{0.641} & \underline{0.469} & \textbf{2157}  %
    \\

    Waymo~\cite{waymo}
    & 12.66 & 0.312 & \underline{0.593} & 2074
    & \underline{13.27} & 0.377 & 0.594 & 2186
    & \textbf{16.31} & \textbf{0.470} & \textbf{0.485} & \textbf{907}  %
    \\

    \midrule

    Average
    & 12.82 & 0.315 & \underline{0.602} & 3499
    & 12.28 & 0.352 & 0.610 & 3745
    & \textbf{17.61} & \textbf{0.528} & \textbf{0.441} & \textbf{1239}  %
    \\
    
    \midrule
    \midrule
    \multicolumn{12}{@{}l }
    {\textbf{\textit{\ourbench}: Zero-shot}} \\
    \midrule
    \midrule

    & \multicolumn{4}{c|}{GCD~\cite{vanhoorick2024gcd}}
    & \multicolumn{4}{c|}{TrajAttn~\cite{xiao2025trajectory}}
    & \multicolumn{4}{c}{GEN3C~\cite{ren2025gen3c}}
    \\

    \cmidrule(lr){2-5}\cmidrule(lr){6-9}\cmidrule(lr){10-13}

    Argoverse~\cite{Argoverse2}
    & \underline{11.45} & \textbf{0.403} & 0.682 & 3536
    & 10.62 & 0.317 & 0.665 & 1717
    & 10.52 & 0.319 & 0.680 & 1620
    \\

    AssemblyHands~\cite{ohkawa:cvpr23}
    & 9.77 & 0.262 & 0.759 & 3956
    & 9.97 & 0.266 & 0.736 & \underline{2643}
    & 9.86 & 0.237 & 0.732 & \textbf{1742}
    \\

    DDAD~\cite{ddad}
    & 9.81 & 0.278 & 0.660 & 5026
    & 9.16 & 0.244 & 0.620 & 1977
    & 9.35 & 0.259 & 0.600 & \underline{1645}
    \\

    DROID~\cite{droid}
    & \underline{11.81} & 0.315 & 0.690 & 5358
    & 10.83 & 0.320 & 0.678 & \underline{3129}
    & 10.56 & 0.276 & 0.674 & 3232
    \\

    EgoExo4D~\cite{egoexo4d}
    & \underline{11.98} & \underline{0.239} & 0.668 & 3862
    & 11.31 & 0.203 & 0.653 & 2576
    & 11.40 & 0.193 & 0.651 & \underline{2210}
    \\

    \midrule

    Average
    & 10.96 & 0.299 & 0.692 & 4348
    & 10.38 & 0.270 & 0.670 & 2408
    & 10.34 & 0.257 & 0.667 & \underline{2090}
    \\

    \midrule
   & \multicolumn{4}{c|}{TrajCrafter~\cite{mark2025trajectorycrafter}}
   & \multicolumn{4}{c|}{CogNVS~\cite{chen2025cognvs}}
   & \multicolumn{4}{c}{Ours (\method)}
    \\
    \cmidrule(lr){2-5}\cmidrule(lr){6-9}\cmidrule(lr){10-13}

    Argoverse~\cite{Argoverse2}
    & 10.67 & 0.325 & \underline{0.621} & 1747
    & 10.76 & 0.360 & \textbf{0.610} & \underline{1469}
    & \textbf{11.87} & \underline{0.388} & 0.645 & \textbf{982}  %
    \\

    AssemblyHands~\cite{ohkawa:cvpr23}
    & \textbf{11.45} & 0.248 & \textbf{0.701} & 4300
    & 9.93 & \textbf{0.281} & \textbf{0.701} & 3862
    & \underline{10.39} & \underline{0.275} & \underline{0.711} & 2870  %
    \\

    DDAD~\cite{ddad}
    & \underline{10.73} & 0.300 & \textbf{0.558} & 2274
    & \textbf{10.81} & \textbf{0.355} & 0.572 & 1879
    & 10.63 & \underline{0.314} & \underline{0.568} & \textbf{1419}  %
    \\

    DROID~\cite{droid}
    & 11.37 & 0.339 & \underline{0.614} & 3395
    & 10.48 & \underline{0.358} & 0.632 & 4312
    & \textbf{11.98} & \textbf{0.399} & \textbf{0.603} & \textbf{1956}  %
    \\

    EgoExo4D~\cite{egoexo4d}
    & 11.27 & 0.180 & \underline{0.647} & 3303
    & 10.52 & 0.227 & 0.683 & 2967
    & \textbf{13.02} & \textbf{0.281} & \textbf{0.578} & \textbf{1555}  %
    \\

    \midrule

    Average
    & \underline{11.10} & 0.279 & \underline{0.628} & 3004
    & 10.50 & \underline{0.316} & 0.640 & 2898
    & \textbf{11.58} & \textbf{0.331} & \textbf{0.621} & \textbf{1757}  %
    \\
    
    \bottomrule
  \end{tabular}
  }

  \vspace{2mm}
  
  \caption{
  \textbf{Extreme DVS results (direct GT comparisons).}
Note that \emph{in-distribution} datasets are part of \method's training mixture, but might be zero-shot for some of the baselines, hence we provide these results for completeness.
For qualitative comparison, please refer to Figure~\ref{fig:teaser}. %
  }
  \label{tab:res_ext_psl}
  
\end{table*}

\begin{figure*}
  \centering
  \includegraphics[width=0.48\linewidth]{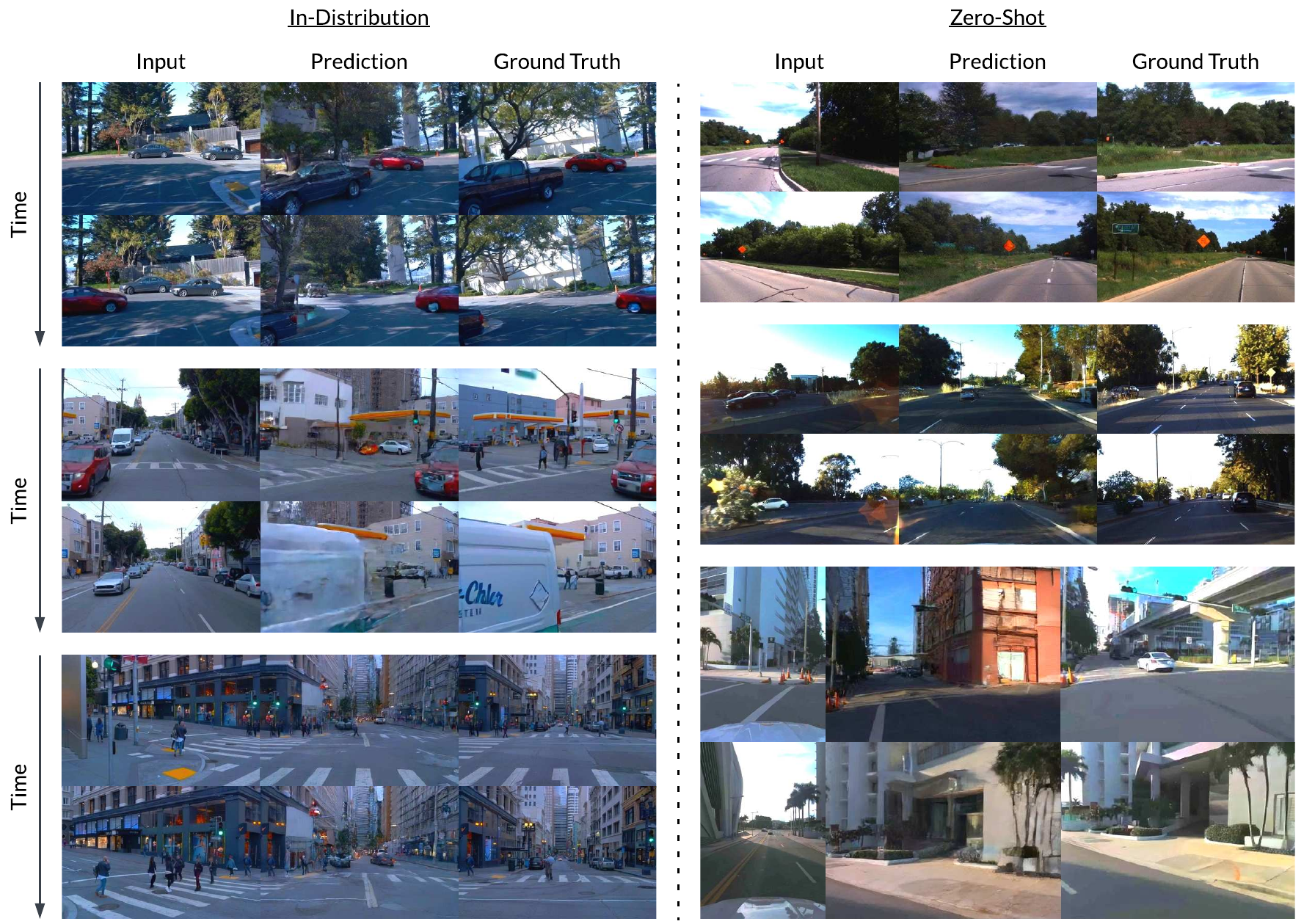}
  \hspace{2mm}
  \includegraphics[width=0.48\linewidth]{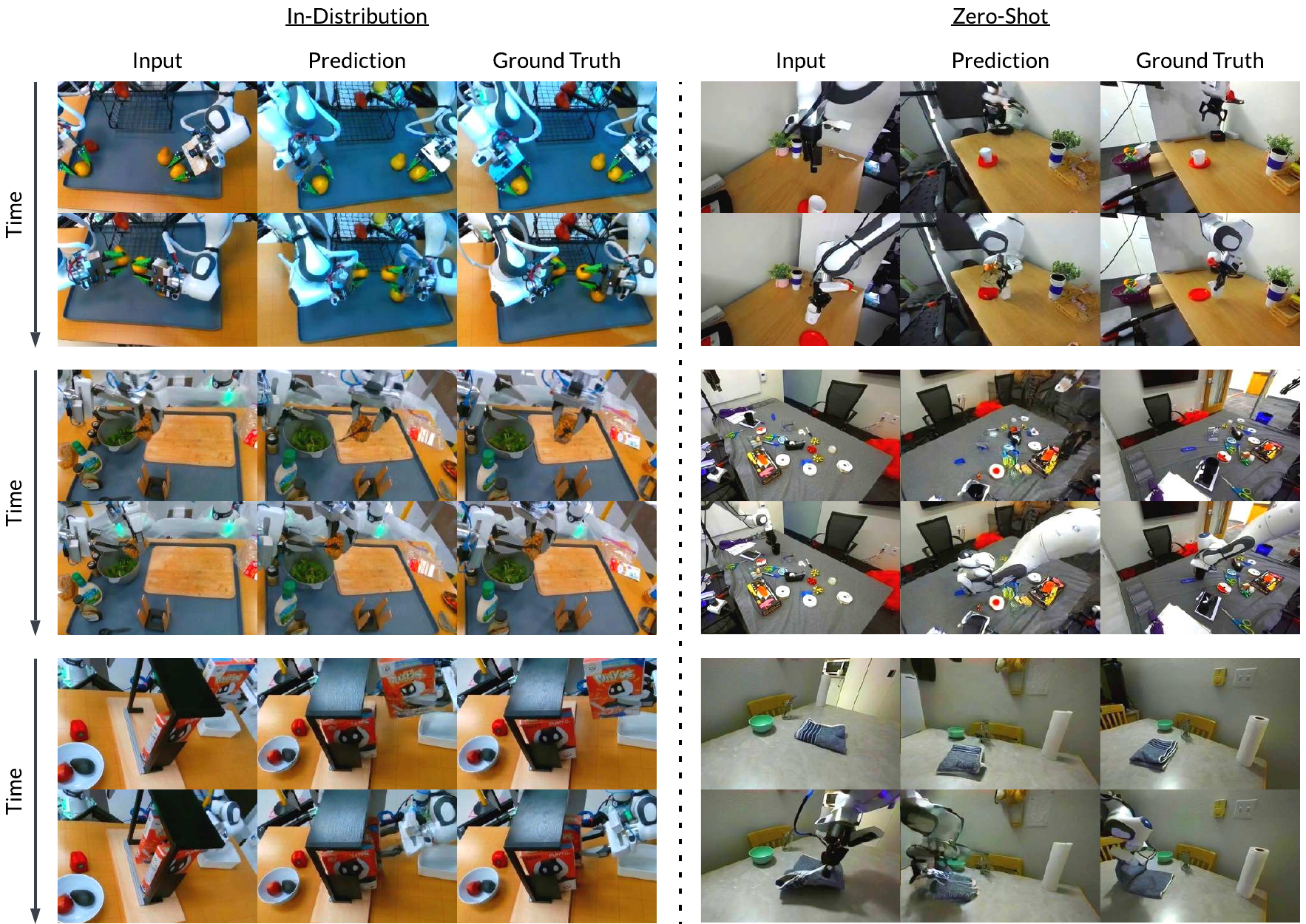}
  \vspace{-0.5mm}
  \caption{
  \textbf{
  \method extreme DVS results}
  on \textbf{driving} (left) and \textbf{robotics} (right) benchmarks.
  We show both in-domain and zero-shot results.
  For driving videos, we focus on the three frontal cameras, whereas for robotics, we focus on all scene cameras.
  }
  \label{fig:res_drive_robot}
\end{figure*}

\begin{figure}[t!]
\vspace{-2.5mm}
  \centering
  \includegraphics[width=0.99\linewidth]{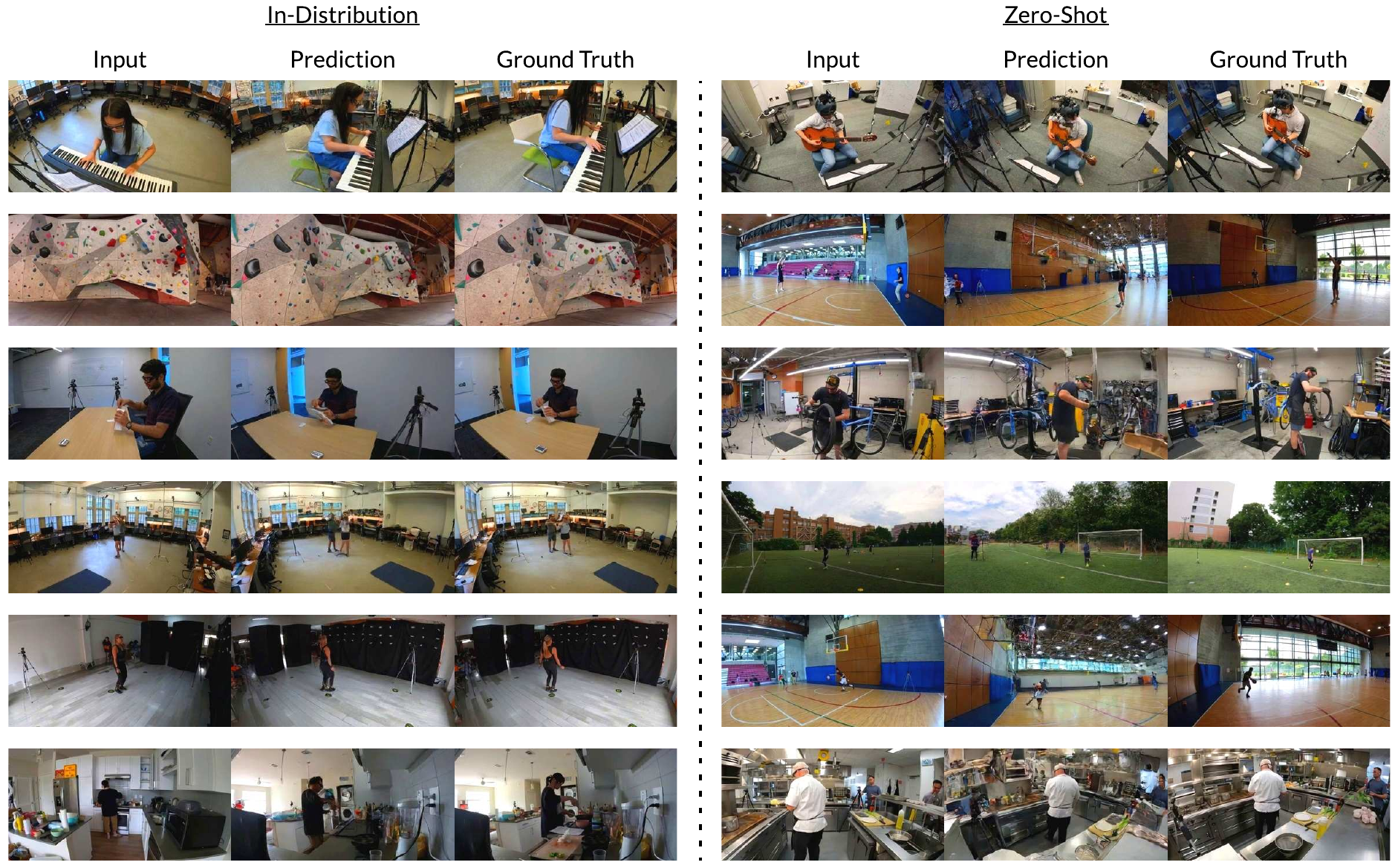}
  \vspace{-0.5mm}
  \caption{
  \textbf{AnyView extreme DVS results on Ego-Exo4D.}
  We show both in-domain and zero-shot results.
  Note that in the zero-shot case, the background often has to be ``guessed'' from the other camera viewpoint, but the inpainted regions (see \eg basketball, soccer) integrate harmoniously with the rest of the scene.
  }
  \vspace{-4mm}
  \label{fig:res_egoexo4d}
\end{figure}

\noindent\textbf{AnyViewBench (extreme DVS).}
We introduce \ourbench, a multi-faceted benchmark that covers datasets across multiple domains (driving, robotics, and human activities), as shown in Table~\ref{tab:data_eval}.
The camera motion patterns range from simple (fixed or linear) to complex (\eg highly non-linear trajectories, changing intrinsics, \etc).
To promote rigorous evaluation, we provide synchronized videos from at least two separate viewpoints for each episode, with well-defined details such as spatial resolution and number of frames, such that ground truth metrics can be calculated in a straightforward manner.

For all \emph{in-distribution} datasets, we separate roughly $10\%$ to serve as validation, and for both \emph{in-distribution} and \emph{zero-shot} datasets, we curate smaller subsets to serve as official test splits.
Moreover, two DROID stations (GuptaLab, ILIAD), as well as certain EgoExo4D institutions (FAIR, NUS) and activities (CPR, Guitar), are held out to serve as \emph{zero-shot} evaluation.
More information about \ourbench can be found in the supplementary material, and it can be downloaded at \href{https://tri-ml.github.io/AnyView/}{tri-ml.github.io/AnyView}.

\definecolor{Hvan}{rgb}{0.0, 0.6, 0.4}
\definecolor{Hlight}{rgb}{0.8, 0.1, 0.7}
\definecolor{Hcar}{rgb}{0.0, 0.3, 0.8}
\definecolor{Hrecon}{rgb}{0.0, 0.6, 0.0}

\begin{figure*}[t]
\vspace{-1.5mm}
  \centering
  \begin{minipage}[t]{0.48\linewidth}
    \centering
    \includegraphics[width=0.98\linewidth]{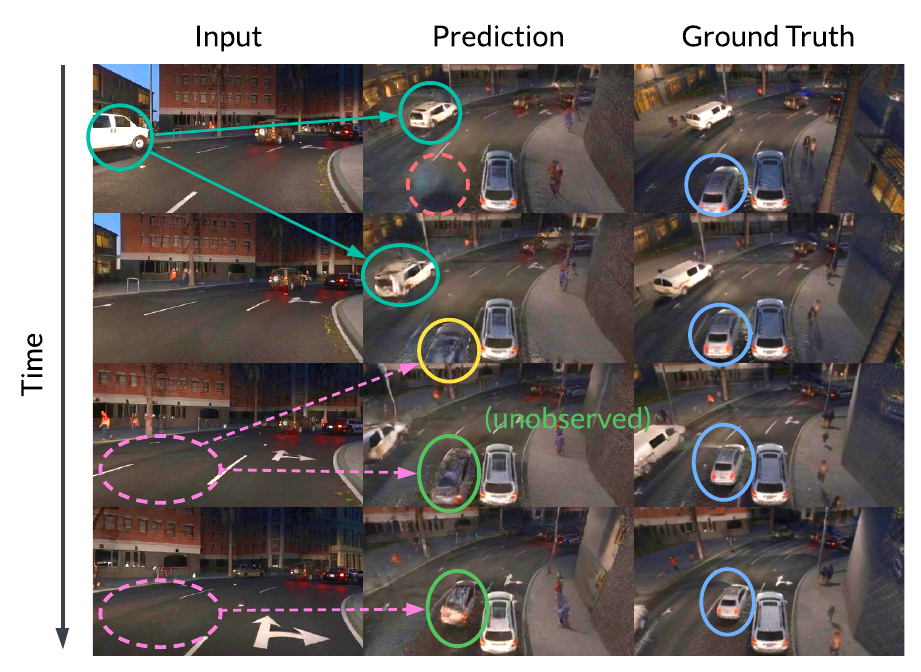}
    \vspace{1mm}

    \raggedright
    \small\textbf{(a)} \textbf{Headlight reflections.}
    While the {\textcolor{Hvan}{white van}} is partially visible in the first few frames, and thus gets depicted accurately in the remainder of the video, the {\textcolor{Hcar}{white car}} is never observed in \emph{any} input frame.
    Instead, \method appears to pick up on the {\textcolor{Hlight}{headlights reflecting on the road}}.
    Although the {\textcolor{Hrecon}{reconstructed car}} does not have the correct appearance, the model estimates its trajectory by tracking the reflection over time.
  \end{minipage}%
  \hfill
  \begin{minipage}[t]{0.48\linewidth}
    \centering
    \includegraphics[width=0.98\linewidth]{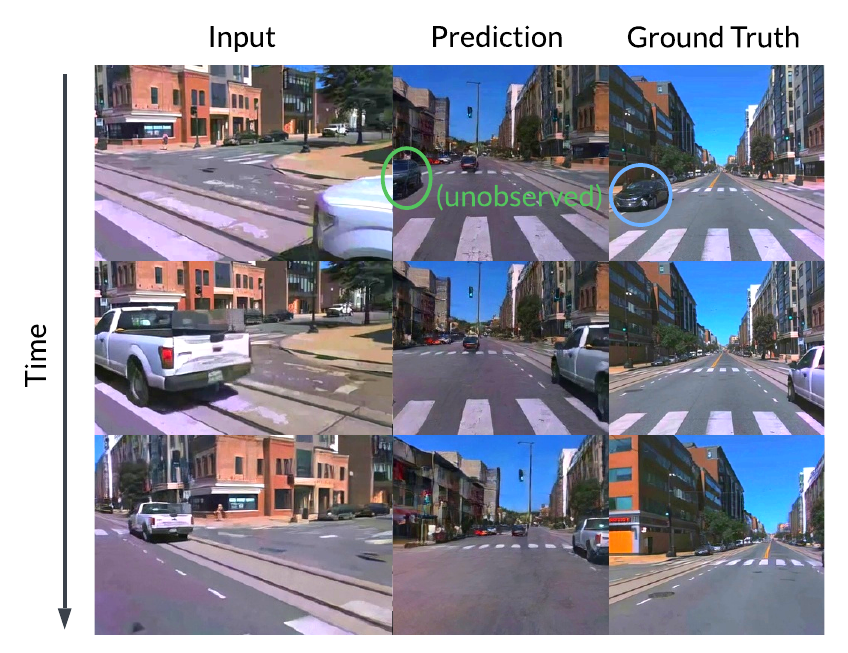}
    \vspace{1mm}

    \raggedright
    \small\textbf{(b)} \textbf{Driving behavior.}
    This zero-shot ArgoVerse scene depicts the ego vehicle pausing for a moment, and then turning left.
    The model correctly hallucinates the \textcolor{Hrecon}{black car} passing by on the left of the generated video, despite never observing it, presumably based on the suspicion that the driver must be waiting at the green light because of oncoming traffic before executing the unprotected left turn.
  \end{minipage}
  \vspace{-0.5mm}
  \caption{
  \textbf{Examples of advanced reasoning within \method}, as a way to indirectly guide generation in unobserved parts of the scene.
  }
  \vspace{-3mm}
  \label{fig:advanced}
\end{figure*}

\subsection{Baselines}

Most current DVS methods face key limitations: the input video must be captured from a strictly \emph{static} camera~\cite{vanhoorick2024gcd}, or from a strictly \emph{dynamic} camera~\cite{ren2025gen3c,chen2025cognvs}, %
or both input and output videos must start from the same position~\cite{xiao2025trajectory,mark2025trajectorycrafter,bai2025recammaster},
or the camera controlling mechanism has limited degrees of freedom~\cite{vanhoorick2024gcd,mark2025trajectorycrafter,bai2025recammaster}.
As a result, methods that excel in certain conditions might be incompatible with slightly different evaluation settings, hindering standardized evaluation across multiple benchmarks. 
More information detailing all prior works we considered as baselines can be found in the supplementary material. 
Most of these models already evaluate on at least a subset of the ``narrow'' benchmarks, but we additionally evaluate them (doing our best effort to project down to and accommodate the space of supported camera transformations as needed)
on \ourbench, which embodies the ``extreme'' benchmarks.
When evaluating baseline methods that require depth estimation to render reprojected images, we use DepthAnythingV2~\cite{depth_anything_v2} and tune the maximum depth parameter for each dataset to achieve the best alignment between reprojected and ground truth images.

\subsection{Results}

\noindent\textbf{Metrics.}
Following standard convention, we report DVS results in terms of PSNR (dB), SSIM, and LPIPS (VGG), averaged over all frames in the generated video,
 as well as FVD~\cite{unterthiner2019fvd}.
Note that pixel-level metrics can only attest to how similar generated predictions are to the ground truth, but not necessarily how realistic and plausible they are when the true underlying scene cannot be fully known due to lack of overlap between viewpoints; FVD complements these by capturing perceptual realism independently of exact pixel correspondence.

\noindent\textbf{Narrow DVS.}
Quantitative results on existing narrow DVS benchmarks are reported in Table~\ref{tab:res_nar_psl}, with qualitative results  in~\cref{fig:res_pardom,fig:res_dycheck}.
For completeness, we also include metrics as reported by other papers, as well as evaluate the baselines ourselves when possible. 
\method outperforms GCD~\cite{vanhoorick2024gcd}, the only baseline that does not require explicit depth estimation or reprojection, by a large margin, and compares favorably with explicit depth reprojection methods --- and those that require expensive test-time-optimization --- in most metrics. This \emph{narrow} setting (i.e., large overlapping regions with small viewpoint changes) is particularly well-suited for such methods, since a lot of information can be directly transferred across viewpoints, and the model is tasked solely with inpainting missing regions.

\noindent\textbf{Extreme DVS.}
Next, we report results in \emph{extreme} DVS setting using \ourbench, with quantitative results in Table~\ref{tab:res_ext_psl}
and illustrations in~\cref{fig:res_drive_robot,fig:res_egoexo4d}.\footnote{FVD follows the official TF-GAN implementation with the I3D network; high values are due to very low sample size per dataset ($20-64$ instead of $\gg 1000$).}
These scenarios are much more challenging, since they require implicit 4D understanding to ensure spatiotemporal consistency. 
For example, in real-world driving, the amount of spatial overlap between neighboring cameras is generally small, meaning that when the model is prompted with generating the front-left view based solely on the front view (or vice-versa), it has to plausibly infer the majority of the scene based on little information.
However, if the ego vehicle is moving, information is able to eventually ``leak'' into other views and propagate across the entire sequence, further limiting the space of ``correct'' generations.

\noindent\textbf{Geometric consistency.}
To assess whether \method's implicit representation captures 3D geometry rather than merely plausible pixels, we estimate camera extrinsics from the \emph{generated} videos with MapAnything~\cite{mapanything} and compare them to ground-truth estimates. As reported in Table~\ref{tab:pose}, \method outperforms all baselines in translation and rotation accuracy by a large margin, in-distribution and zero-shot, despite using no explicit geometric supervision, unlike several baselines (\eg GEN3C, which relies on DepthAnythingV2~\cite{depth_anything_v2}). This indicates a genuine, implicit understanding of 3D structure.

\begin{table}[t]
  \centering
  \smaller
  \setlength{\tabcolsep}{4pt}
  \renewcommand{\arraystretch}{0.99}
  \begin{tabular}{@{}l|ccc|ccc@{}}
    \toprule
    \multirow{2}{*}{Method}
    & \multicolumn{3}{c|}{In-distribution}
    & \multicolumn{3}{c}{Zero-shot}
    \\
    \cmidrule(lr){2-4}\cmidrule(lr){5-7}
    & Transl.$\downarrow$ & Rot.$\downarrow$ & AUC$_{15}$$\uparrow$
    & Transl.$\downarrow$ & Rot.$\downarrow$ & AUC$_{15}$$\uparrow$
    \\
    \midrule
    GCD~\cite{vanhoorick2024gcd}
    & \underline{0.919} & 2.774 & 19.87
    & \underline{1.201} & \underline{2.381} & 12.14 \\
    TrajAttn~\cite{xiao2025trajectory}
    & 2.219 & 5.872 & 11.89
    & 2.376 & 5.156 & 8.01 \\
    GEN3C~\cite{ren2025gen3c}
    & 1.798 & 5.388 & 13.13
    & 2.234 & 4.792 & 7.76 \\
    TrajCrafter~\cite{mark2025trajectorycrafter}
    & 1.021 & 3.101 & \underline{21.30}
    & 1.421 & 2.571 & 11.79 \\
    CogNVS~\cite{chen2025cognvs}
    & 1.215 & \underline{2.490} & 17.42
    & 1.398 & 2.759 & \underline{12.31} \\
    \midrule
    Ours (\method)
    & \textbf{0.217} & \textbf{0.654} & \textbf{43.38}
    & \textbf{0.608} & \textbf{1.781} & \textbf{16.52} \\
    \bottomrule
  \end{tabular}

  \vspace{2mm}
  \caption{
  \textbf{Camera pose estimation} results on \ourbench.
  \emph{Transl.} and \emph{Rot.} are the translation and rotation (degrees) errors between camera extrinsics estimated from the generated versus ground-truth videos; \emph{AUC$_{15}$} is the area under the pose-accuracy curve at a binning threshold of 15.
  }
  \label{tab:pose}
  
\end{table}

\begin{table*}[t]
  \centering
  \smaller
  \setlength{\tabcolsep}{2pt}
  \renewcommand{\arraystretch}{0.99}

  \begin{tabular}{@{}l |ccc|ccc|ccc@{}}
    \toprule
    \multirow{2}{*}{Dataset}
    & \multicolumn{3}{c|}{From scratch}
    & \multicolumn{3}{c|}{GCD data}
    & \multicolumn{3}{c}{Ours (\method)}
    \\
    \cmidrule(lr){2-4}\cmidrule(lr){5-7}\cmidrule(lr){8-10}
    & PSNR$\uparrow$ & SSIM$\uparrow$ & LPIPS$\downarrow$
    & PSNR$\uparrow$ & SSIM$\uparrow$ & LPIPS$\downarrow$
    & PSNR$\uparrow$ & SSIM$\uparrow$ & LPIPS$\downarrow$
    \\

    \midrule
    \midrule
    \multicolumn{10}{@{}l}{\textbf{\textit{\ourbench}: In-distribution}} \\
    \midrule
    \midrule
        DROID & \textbf{16.55} & \textbf{0.520} & \textbf{0.414} & 10.70 & 0.254 & 0.680 & \underline{14.42} & \underline{0.443} & \underline{0.473} \\  %
        EgoExo4D & \underline{17.11} & \underline{0.464} & \underline{0.450} & 12.37 & 0.253 & 0.680 & \textbf{17.46} & \textbf{0.493} & \textbf{0.410} \\  %
        \LBM & \textbf{20.15} & \textbf{0.711} & \textbf{0.285} & 12.52 & 0.366 & 0.652 & \underline{17.63} & \underline{0.619} & \underline{0.386} \\  %
        Kubric-4D & 16.50 & 0.393 & 0.535 & \underline{19.87} & \underline{0.508} & \textbf{0.410} & \textbf{19.91} & \textbf{0.541} & \underline{0.427} \\  %
        Kubric-5D & 14.77 & \underline{0.382} & \underline{0.570} & \underline{14.84} & 0.367 & 0.608 & \textbf{17.04} & \textbf{0.460} & \textbf{0.488} \\  %
        Lyft & \underline{12.01} & \underline{0.432} & \underline{0.481} & 8.72 & 0.267 & 0.691 & \textbf{15.29} & \textbf{0.560} & \textbf{0.389} \\  %
        ParDom-4D & 21.60 & 0.561 & 0.507 & \textbf{24.23} & \textbf{0.698} & \textbf{0.378} & \underline{22.85} & \underline{0.641} & \underline{0.469} \\  %
        Waymo & \underline{13.68} & \underline{0.426} & \underline{0.550} & 13.05 & 0.361 & 0.629 & \textbf{16.31} & \textbf{0.470} & \textbf{0.485} \\  %
    \midrule
        Average & \underline{16.55} & \underline{0.486} & \underline{0.474} & 14.54 & 0.384 & 0.591 & \textbf{17.61} & \textbf{0.528} & \textbf{0.441} \\  %

    \midrule
    \midrule
    \multicolumn{10}{@{}l}{\textbf{\textit{\ourbench}: Zero-shot}} \\
    \midrule
    \midrule
        Argoverse & \underline{11.46} & 0.327 & \underline{0.660} & 11.06 & \underline{0.343} & 0.670 & \textbf{11.87} & \textbf{0.388} & \textbf{0.645} \\  %
        AssemblyHands & \textbf{10.66} & 0.220 & \underline{0.713} & \underline{10.58} & \underline{0.267} & 0.739 & 10.39 & \textbf{0.275} & \textbf{0.711} \\  %
        DDAD & \underline{10.28} & \underline{0.299} & \underline{0.587} & 9.53 & 0.254 & 0.635 & \textbf{10.63} & \textbf{0.314} & \textbf{0.568} \\  %
        DROID & \underline{11.63} & \underline{0.329} & \underline{0.617} & 11.53 & 0.327 & 0.678 & \textbf{11.98} & \textbf{0.399} & \textbf{0.603} \\  %
        EgoExo4D & \underline{12.74} & 0.206 & \underline{0.635} & 12.26 & \underline{0.225} & 0.680 & \textbf{13.02} & \textbf{0.281} & \textbf{0.578} \\  %
    \midrule
        Average & \underline{11.35} & 0.276 & \underline{0.642} & 10.99 & \underline{0.283} & 0.680 & \textbf{11.58} & \textbf{0.331} & \textbf{0.621} \\  %
    \bottomrule
  \end{tabular}
  
  \vspace{2mm}
  \caption{
  \textbf{Extreme DVS ablations.}
  We compare \method with two ablations of the same model: one trained from scratch (\ie without Cosmos pretrained weights),
  and another finetuned with GCD training data only (\ie Kubric-4D and ParallelDomain-4D)~\cite{vanhoorick2024gcd}.
  All ablations use the same underlying architecture.
  }
  \label{tab:res_abl}
  
\end{table*}

\noindent\textbf{Ablations.}
We also ablate two key design choices of \method in Tables~\ref{tab:res_nar_psl} and~\ref{tab:res_abl}:
(1) training from scratch, as opposed to initializing from Cosmos pretrained weights;
and (2) match the GCD training data by only training on Kubric-4D and ParallelDomain-4D~\cite{vanhoorick2024gcd}, as opposed to using our full \method data mixture as in Figure~\ref{fig:data}.
Training from scratch consistently degrades performance across both narrow and extreme settings, confirming the importance of leveraging a strong video foundation model.
Using only GCD training data yields competitive results on Kubric-4D and ParallelDomain-4D, presumably because they are in-domain, but generalizes poorly to other domains, particularly on DyCheck iPhone.
The full \method training mixture, combining diverse real-world and synthetic data, provides the best overall performance and strongest zero-shot generalization.

\noindent\textbf{Analysis.}
Remarkably, in the upper left scenario in Figure~\ref{fig:res_drive_robot}, the red car arriving at the intersection is predicted on the left view \emph{before} it is visible in the input front view, showing that \method has learned to maintain spatiotemporal consistency, leading to improved performance in areas that otherwise would be ill-defined. 
A related behavior is also observed in the left examples of Figure~\ref{fig:res_egoexo4d}, where \method leverages its foundational knowledge to infer how a basketball court or soccer field should look like from different perspectives. 
Moreover, in~\cref{fig:advanced} we show anecdotal examples of \method leveraging subtle visuals cues to improve generation accuracy in unobserved areas, as evidence of advanced common sense and spatiotemporal reasoning.

Implicitly learning these useful spatiotemporal properties in a data-driven way enables \method to produce more realistic and physically plausible representations of real-world scenarios compared to all baselines. As shown in \cref{fig:teaser}, while methods that rely on potentially inaccurate depth reprojection (\eg TrajAttn and GEN3C) struggle when starting from target poses away from input poses, \method successfully generates smooth, consistent target scenes regardless of camera positioning. Similarly, \method is able to accurately outpaint much larger unobserved portions of the scene compared to methods trained mostly for limited inpainting (\eg TrajCrafter and CogNVS). 
As a consequence of these useful properties, we achieve state of the art zero-shot DVS performance on \ourbench, outperforming all other baseline methods by a significant margin across all considered datasets.

\vspace{-1mm}
\section{Discussion}
\vspace{-1mm}
\label{sec:disc}

In this paper, we propose \emph{\method}, a generalist dynamic view synthesis framework targeting extreme camera displacements.
We also contribute \emph{\ourbench}, a well-rounded benchmark that focuses on highly challenging scenarios from various domains, showing that \method significantly outperforms baselines in such settings with large camera displacement and limited overlap between views.
We remark that \method is not without limitations: it requires known input camera poses, offers no hard geometric guarantees, and can still produce plausible yet incorrect content in heavily occluded regions, particularly under the most extreme viewpoint changes.
Regardless, we hope that this work provides a useful building block towards improving video foundation models and 4D representations, with potential applications in dynamic scene reconstruction, world models, robotics, self-driving, and more.

\bibliographystyle{splncs04}
\bibliography{main}

\clearpage

{
\centering
\Large
\textbf{AnyView: \\
Synthesizing Any Novel View in Dynamic Scenes} \\
\vspace{0.5em}Supplementary Material \\
\vspace{1.0em}
}

\appendix

\definecolor{Hteal}{rgb}{0.0, 0.6, 0.4}
\definecolor{Hpink}{rgb}{0.8, 0.1, 0.7}
\definecolor{Hblue}{rgb}{0.0, 0.3, 0.8}
\definecolor{Hgreen}{rgb}{0.0, 0.6, 0.0}
\definecolor{Hred}{rgb}{0.9, 0.2, 0.2}
\definecolor{Hyellow}{rgb}{0.6, 0.6, 0.0}

\begin{figure*}
  \centering
  \includegraphics[width=0.9\linewidth]{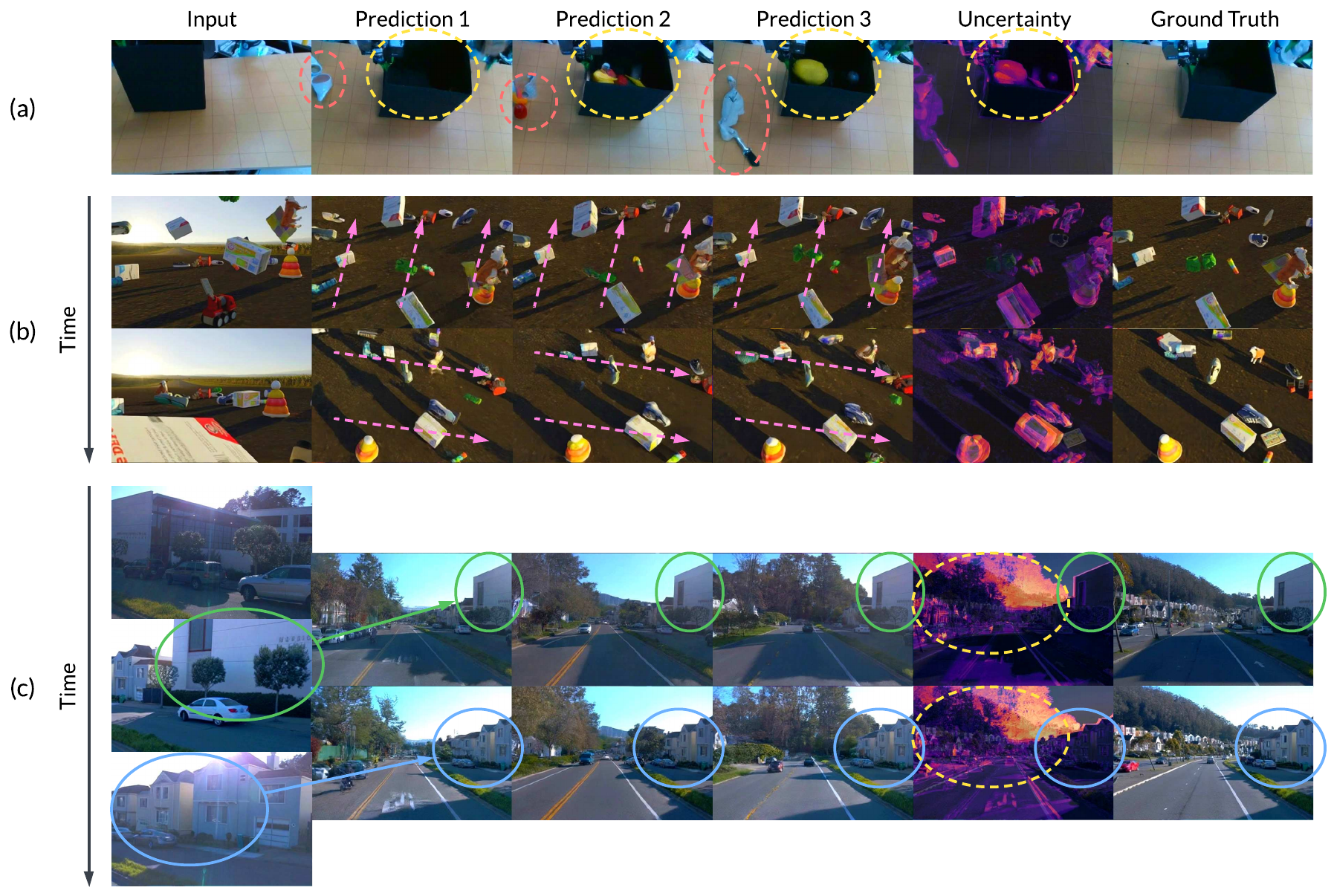}
  \caption{
    \textbf{Uncertainty analysis.}
    In \textbf{(a)}, the model cannot see what is contained {\color{Hyellow}inside the black bin} because the contents are occluded, and resorts to predicting fruit (since those objects are common in \LBM), in addition to spawning {\color{Hred}spurious objects} out-of-frame on the left.
    In \textbf{(b)}, we mainly observe variations of object positions along the {\color{Hpink}{input viewing direction}} (overlayed with pink arrows for clarity), which presumably stems primarily from uncertainty in terms of implicit depth estimation that the model has to perform internally as part of the representation.
    In \textbf{(c)}, only the front-right view is seen, which passes by {\color{Hgreen}several} {\color{Hblue}buildings} that are reconstructed correctly in all samples (= front view).
    Meanwhile, the {\color{Hyellow}left half} of these output videos has more diversity since it is never directly observed.
  }
  \label{fig:uncertain}
\end{figure*}

\section{Uncertainty Analysis}

Figure~\ref{fig:uncertain} showcases how \method represents and expresses uncertainty. We calculate this by running the diffusion model multiple times to collect independent samples from the conditional distribution, and plotting the per-pixel diversity between these predictions as a spatial heatmap.
Each generation is conditioned on the same input signals, and represents a possible version of what the other viewpoint \emph{could} look like.
Even if these outputs are not technically correct, due to the inherent ambiguity of the task at hand, they are still reasonable, realistic, and self-consistent, demonstrating that \method learns a powerful probabilistic representation that encodes the natural multimodality of unobserved parts of the world.

\definecolor{Hgreen}{rgb}{0.0, 0.6, 0.0}
\definecolor{Hred}{rgb}{0.9, 0.1, 0.1}

\begin{figure*}
  \centering
  \includegraphics[width=0.9\linewidth]{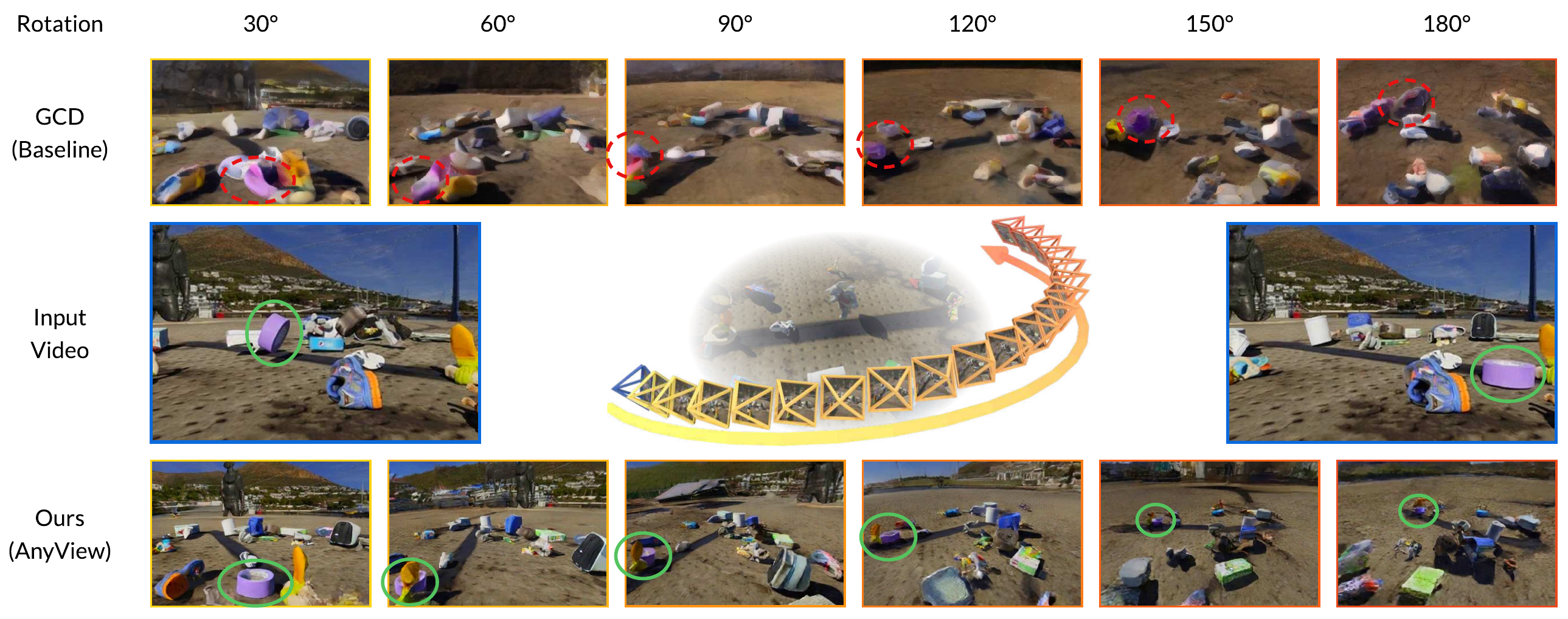}
  \caption{
    \textbf{Gradually increasing target azimuth.}
    As we increase the difficulty of the task by rotating the virtual camera over larger and larger angles away from the observed camera in this Kubric scene, GCD~\cite{vanhoorick2024gcd} produces garbled outputs where objects become {\color{Hred}essentially unrecognizable}.
    In contrast, \method maintains {\color{Hgreen}clear spatiotemporal correspondence} across dramatic viewpoint changes, demonstrating significantly enhanced 4D understanding over previous methods.
  }
  \label{fig:sweep}
\end{figure*}

\section{Additional Qualitative Results}
\label{sec:more_results}

We complement the qualitative results depicted in the main paper with the following:

    {Figure~\ref{fig:sweep}} compares the performance of \method against GCD~\cite{vanhoorick2024gcd} over increasingly wide horizontal camera displacements, showing that \method maintains better spatio-temporal consistency over large viewpoint changes.

\begin{figure*}
  \centering
  \includegraphics[width=0.7\linewidth]{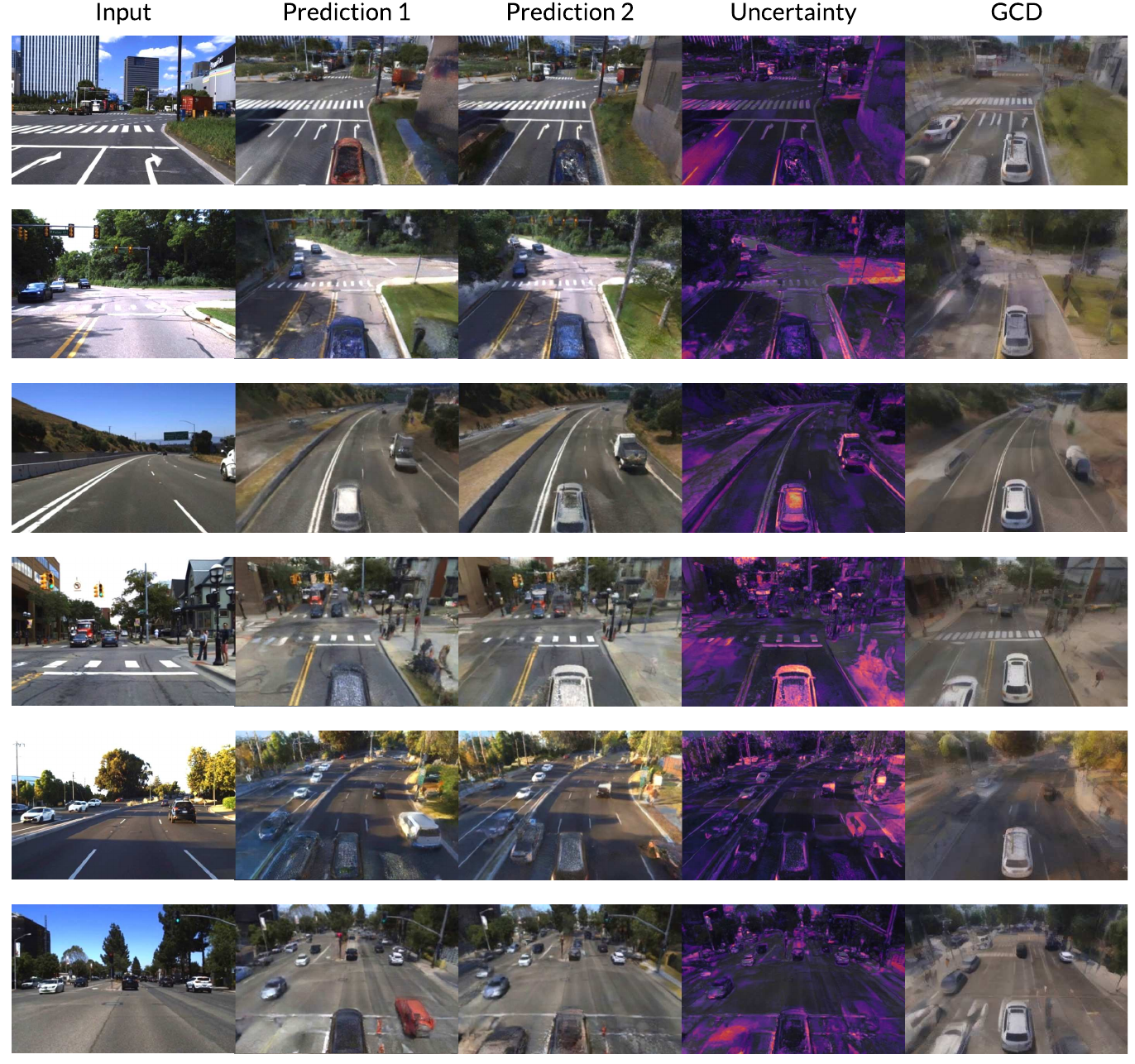}
  \caption{
    \textbf{Upward view synthesis on real-world driving scenarios.}
    We compare \method with GCD~\cite{vanhoorick2024gcd} on DDAD~\cite{ddad}, which is a zero-shot dataset for both methods.
    \method generates much clearer predictions: almost every car that the model can see is reconstructed with high fidelity and accurate dynamics, whereas GCD often suffers from blurry artefacts, which worsen the further away one looks from the ego vehicle.
  }
  \label{fig:eed}
\end{figure*}

    {Figure~\ref{fig:eed}} shows top-down view synthesis on real-world (\emph{DDAD}) driving scenes, where we also compare against the GCD baseline. This effectively tests each model's sim-to-real trajectory generalization capability, since the only training videos corresponding to similar viewpoint configurations (albeit still not the same) come from synthetic data (\emph{ParallelDomain}).

Moreover, we include all figures present in the paper as videos in the project webpage: \href{https://tri-ml.github.io/AnyView/}{tri-ml.github.io/AnyView}.
We highly encourage the reader to browse these results, since it is difficult otherwise to communicate 4D results through 2D PDF files.

\section{Training Datasets}
\label{sec:data_train}

\begin{table*}
  \centering
  \tiny

\scalebox{0.97}{%
  \begin{tabular}{@{}lccccccc@{}}
    \toprule
    Dataset & S/R & Domain & Type & \# Cameras
    & \# Episodes
    & Resolution & Weight
    \\
    \midrule
    
    DL3DV-10K~\cite{ling2024dl3dv} & Real & Indoor + Outdoor & 3D & --
    & 5,906 & 41 frames @ $384 \times 208$ & 6.3~\%
    \\
    
    DROID~\cite{droid} & Real & Robotics & 4D & 2 (exo only)
    & 29,712 & 29 frames @ $384 \times 208$ & 12.5~\%
    \\
    
    Ego-Exo4D~\cite{egoexo4d} & Real & Human Activity & 4D & 4 -- 5 (exo only)
    & 2,489 & 41 frames @ $384 \times 208$ & 9.4~\%
    \\
    
    \LBM & Sim + Real & Robotics & 4D & 2 (exo only)
    & 53,886 & 41 frames @ $336 \times 256$ & 12.5~\%
    \\
    
    Kubric~\cite{greff2021kubric} & Sim & Multi-Object & 4D & 16 (exo only)
    & 12,400 & 41 frames @ $384 \times 256$ & 15.6~\%
    \\
    
    Lyft~\cite{lyftl5} & Real & Driving & 4D & 6 (ego only)
    & 296 & 41 frames @ $384 \times 320$ & 3.1~\%
    \\

    ParallelDomain~\cite{parallel_domain} & Sim & Driving & 4D & 19 (16 exo + 3 ego)
    & 7,352 & 41 frames @ $384 \times 256$ & 18.8~\%
    \\

    RealEstate-10K~\cite{re10k} & Real & Indoor + Outdoor & 3D & --
    & 34,968 & 41 frames @ $384 \times 208$ & 6.3~\%
    \\
    
    ScanNet~\cite{dai2017scannet} & Real & Indoor & 3D & --
    & 1,357 & 41 frames @ $384 \times 288$ & 3.1~\%
    \\
    
    TartanAir~\cite{wang2020tartanair} & Sim & Indoor + Outdoor & 3D & --
    & 369 & 41 frames @ $384 \times 288$ & 3.1~\%
    \\
    
    Waymo~\cite{waymo} & Real & Driving & 4D & 5 (ego only)
    & 798 & 41 frames @ $384 \times \{176,256\}$ & 3.1~\%
    \\
    
    WildRGB-D~\cite{xia2024rgbd} & Real & Single-Object & 3D & --
    & 23,002 & 41 frames @ $384 \times 288$ & 6.3~\%
    \\
    
    \bottomrule
  \end{tabular}
}
  
  \vspace{2mm}
  
  \caption{
  \textbf{\method training datasets.}
  We use a weighted mixture of both static and dynamic data sources that combines multiple domains of interest.
  For multi-view video (4D) datasets, if there are more than two cameras, we randomly sample an input + ground truth pair for each training sample.
  For static (3D) datasets, with videos typically consisting of only one moving camera, %
  we randomly sample subclips and treat them as different cameras for the purposes of training and evaluation.
  }
  \label{tab:data_train}
  
\end{table*}

Here, we provide additional details about the \method training mixture, also summarized in Table~\ref{tab:data_train}. For all training datasets, we randomly selected around 10\% of sequences to serve as \emph{in-distribution} validation, from which many of the official \ourbench test splits were curated.

\begin{itemize}
    \item \textbf{Driving:}
    Most autonomous driving rigs have a set of well-calibrated RGB cameras mounted around the vehicle, providing plenty of real-world, \emph{egocentric} (outward-facing), temporally synchronized video footage.
    We additionally capitalize on synthetic data to provide \emph{exocentric} (inward-facing) viewpoints that otherwise do not naturally occur in such datasets.
    For training, we use the \emph{Woven Planet (Lyft) Level 5}~\cite{lyftl5}, \emph{ParallelDomain}~\cite{vanhoorick2024gcd,guda,draft}, and \emph{Waymo Open} (Perception)~\cite{waymo} datasets.
    
    \vspace{0.1cm}
    \item \textbf{Robotics:}
    To enable our model to operate in embodied AI contexts, we use \emph{DROID}~\cite{droid} with the improved calibration parameters provided in~\cite{irshad2024scalingupcalibration}.
    This dataset was captured at many locations around the world, and laboratories tend to have significantly different appearance, lighting, camera positions, and calibration quality.
    We also include a large collection of internally recorded bimanual and single-arm tabletop robotics demonstrations, denoted \emph{\LBM}.
    
    \vspace{0.1cm}
    \item \textbf{3D:}
    Because multi-view video is expensive to collect and therefore rather small in overall scale, we leverage single-view, posed videos of static scenes as an additional data source. Following~\cite{ren2025gen3c,mark2025trajectorycrafter}, we adopt \emph{DL3DV-10K}~\cite{ling2024dl3dv} and \emph{RealEstate-10K}~\cite{re10k}.
    We also include \emph{ScanNet}~\cite{dai2017scannet}, \emph{TartanAir}~\cite{wang2020tartanair}, and \emph{WildRGB-D}~\cite{xia2024rgbd}.
    Because these environments are not dynamic, each frame can essentially be handled as if it were an independent camera, without any inherent temporal ordering.
    We randomly sample non-overlapping segments of 41 frames at training time, and treat them as two separate viewpoints.
    
    \vspace{0.1cm}
    \item \textbf{Other:}
    This catch-all category covers all remaining multi-view video datasets, including \emph{Kubric-4D}~\cite{vanhoorick2024gcd} and \emph{Kubric-5D}~\cite{greff2021kubric} with synthetic multi-object interaction and physics, as well as \ie \emph{Ego-Exo4D}~\cite{egoexo4d}, depicting complex human activities in cluttered scenes.

\end{itemize}

\begin{figure*}
  \centering
  \includegraphics[width=1\linewidth]{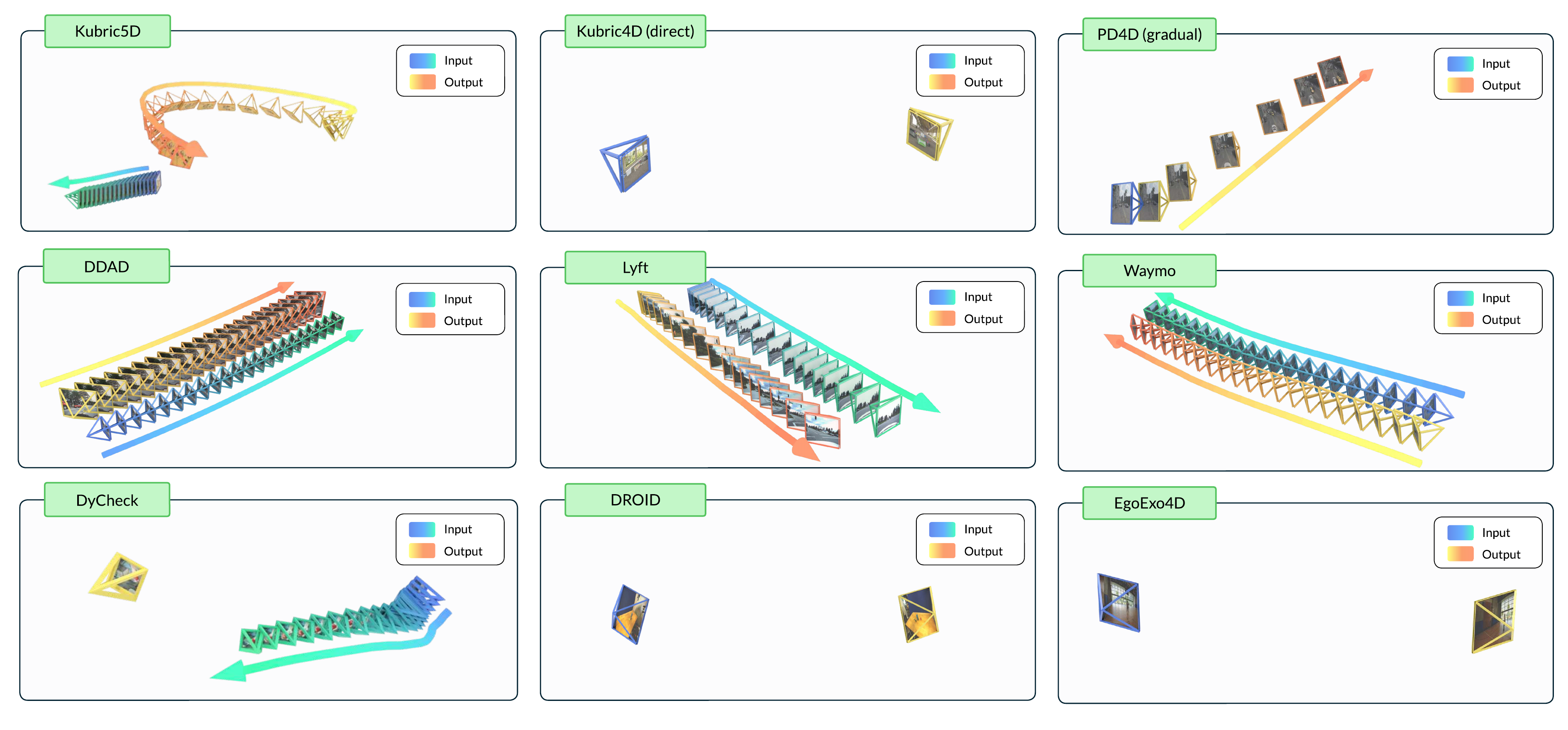}
  \caption{
    \textbf{Diversity of camera trajectories.}
    Samples of dataset camera trajectories illustrating the diversity of motion patterns used in our evaluation.
  }
  \label{fig:trajectories}
\end{figure*}

In Figure~\ref{fig:trajectories}, we provide additional examples of input and target camera poses of various episodes across training and evaluation sets to illustrate the diversity.

\definecolor{Hgreen}{rgb}{0.0, 0.6, 0.0}
\definecolor{Hred}{rgb}{0.9, 0.1, 0.1}

\subsection{Kubric-5D}
\emph{Kubric-5D} is our newly introduced extension of Kubric-4D, rendered with substantially more complex camera configurations and object placements; the ``5D'' name reflects the additional camera variation (dynamic trajectories and varying intrinsics) layered on top of the underlying 4D scene. Whereas Kubric-4D uses static cameras with a constant focal length, facing a small cluster of free-falling objects, \emph{Kubric-5D} introduces dynamic cameras with varying focal lengths and varying object density, enriching the spatiotemporal information available to the model. Specifically, we rendered $10{,}000$ randomized scenes, each containing $16$ cameras spawned at locations evenly distributed around the world center, with each camera's trajectory type sampled independently. For the focal length, with probability $1/3$ all $16$ cameras in a scene share a preset value, with probability $1/3$ they share a randomly sampled value, and with probability $1/3$ each camera samples its own. Combining a geometry such as \texttt{spiral}, \texttt{radial}, \texttt{line}, \texttt{lissajous}, \etc, with the camera's viewing direction yields $16$ distinct trajectory types (including static). The number of objects and the spawn area are likewise randomized per scene, ranging from dense clusters to sparse scatterings. All videos are rendered at $576 \times 384$ resolution at $24$ FPS for $60$ frames, using the Kubric engine~\cite{greff2021kubric} and code adapted from~\cite{vanhoorick2024gcd}.
The generation code is available at \href{https://github.com/TRI-ML/Kubric-5D}{github.com/TRI-ML/Kubric-5D}, and the rendered dataset, like \ourbench, can be downloaded from our project webpage.

\section{Evaluation Datasets}
\label{sec:data_eval}

Here, we describe the logic of which datasets and subsets are held out for evaluation purposes.

\begin{itemize}
    \item \textbf{Driving:}
    The training sets for \emph{Lyft} and \emph{Waymo} are both recorded exclusively in the United States~\cite{lyftl5,waymo}.
    We hold out \emph{Argoverse}, also recorded in the USA~\cite{Argoverse2} (albeit in mostly non-overlapping cities), because it has portrait videos as the front camera, which do not exist during training.
    We also hold out \emph{DDAD}, because it contains videos recorded in Japan~\cite{ddad}.
    
    \vspace{0.1cm}
    \item \textbf{Robotics:}
    While episodes in \LBM are recorded across multiple stations in both simulation and the real world, \emph{DROID}~\cite{droid} has more visual diversity. We decide to hold out all videos belonging to 2 out of 13 institutions (Gupta Lab, ILIAD) for zero-shot testing.
    
    \vspace{0.1cm}
    \item \textbf{Human Activity:}
    One natural choice for this category is \emph{Ego-Exo4D}~\cite{egoexo4d}, which has highly challenging, real-world scenes, often involving multiple humans, recorded by 4 to 5 inward-facing cameras. We hold out two \emph{institutions} (FAIR, NUS), two \emph{activities} (cpr, guitar), and three \emph{institution-activity pairs} (basketball at Uniandes, piano at Indiana, soccer at UTokyo). Notably, \emph{cpr at NUS} becomes the ``most zero-shot'' combination since both the activity and institution are entirely unseen. Since the cameras used to collect the dataset have noticeable distortion, we implement a non-pinhole camera model to generate the actual viewing rays when given a grid, based on the official code examples that undistort the frames using coefficients stored in each sample.
    We further evaluate on videos from the eight exocentric cameras of the \emph{AssemblyHands}~\cite{ohkawa:cvpr23} dataset, a subset of \emph{Assembly101}~\cite{sener2022assembly101} that has calibrated camera intrinsics and extrinsics.
    The dataset records dexterous hand-object interactions during the assembly and disassembly of pull-apart toys, providing a challenging zero-shot test setting for \method.
    
\end{itemize}

\section{Baselines}
\label{sec:supp_base}

\begin{table*}
  \centering
  \setlength{\tabcolsep}{2pt}
    \tiny

\scalebox{0.78}{%
  \begin{tabular}{lcccccccc}
    \toprule
    Method & Base Model & Training Datasets
    & Resolution
    & Input Cam.
    & Align Start
    & \# DOF
    \\
    \midrule

    GCD~\cite{vanhoorick2024gcd} (2024)
      & SVD-1B~\cite{blattmann2023stable}
      & Kubric-4D, ParDom-4D
      & 14 frames @ $384 \times 256$
      & Either
      & Either
      & 3
      \\

    TrajAttn~\cite{xiao2025trajectory} (2024)
      & SVD-1B~\cite{blattmann2023stable}
      & MiraData
      & 25 frames @ $1024 \times 576$
      & Flexible
      & Yes
      & $6 \cdot T$
      \\

    GEN3C~\cite{ren2025gen3c} (2025)
      & GEN3C-Cosmos-7B
      & Kubric-4D, DL3DV, RE-10K, Waymo OD
      & 121 frames @ $1280 \times 704$
      & Moving
      & Yes
      & $6 \cdot T$
      \\

    TrajCrafter~\cite{mark2025trajectorycrafter} (2025)
      & CogVideoX-Fun-5B~\cite{yang2024cogvideox}
      & OpenVid-1M, DL3DV, RE-10K
      & 49 frames @ $672 \times 384$
      & Flexible
      & Yes\tstar  %
      & 5 %
      \\

    ReCamMaster~\cite{bai2025recammaster} (2025)
      & Wan2.1~\cite{wan2025wan}  %
      & MultiCamVideo
      & 81 frames @ $672 \times 384$
      & Flexible
      & Yes
      & $<1$
      \\

    CogNVS~\cite{chen2025cognvs} (2025)
      & CogVideoX-5B-I2V~\cite{yang2024cogvideox}
      & SA-V, TAO, YT-VOS, DAVIS
      & 49 frames @ $720 \times 480$
      & Moving
      & Flexible
      & 6
      \\

      \midrule

    Ours (\method)
  & Cosmos-2B~\cite{nvidia2025worldsimulationvideofoundation}
  & See Table~\ref{tab:data_train}
  & $[9,41]$ frames @ $576 \times [304,384]$
  & Flexible
  & Flexible
  & $6 \cdot T$
  \\

    \bottomrule
  \end{tabular}
  }
  \vspace{2mm}
  
  \caption{
  \textbf{Description of baselines.}
  Some methods are self-supervised~\cite{xiao2025trajectory,mark2025trajectorycrafter,chen2025cognvs} and/or training-free~\cite{yesiltepe2025dynamicviewsynthesisinverse}, and hence do not require multi-view video datasets for training.
  \emph{Input Cam.} refers to what kind of video a model can accept as input. \emph{Align Start} specifies whether the output trajectory needs to start at the same initial frame, in which case we typically apply the smooth interpolation procedure.
  See Section~\ref{sec:supp_base} for more information.
  {\smaller
  \tstar TrajCrafter is trained with aligned start, but the official implementation does include limited support for non-aligned starting point inference.
  }
  }
  \label{tab:baselines}
  
\end{table*}

Here, we outline how each baseline was adapted to \ourbench.
In each case, when a method predicts \emph{fewer} frames than the evaluation episode, we run the model multiple times in a sliding window fashion until the full video is covered, and average metrics such that each frame is used exactly once.
In the opposite scenario, \ie when a method predicts \emph{more} frames than necessary, we simply discard the superfluous ones.

We provide the evaluated methods with ground truth camera pose and intrinsics, and when a method needs depth we use DepthAnythingV2 \cite{depth_anything_v2} to calculate metric depths maps since the ground truth pose we use are in metric space.

Some methods are trained to operate with smooth camera trajectories,
and their performance degrades when there is minimal overlap between the target and input trajectories in the beginning of the videos.
However, many trajectories in \ourbench exhibit precisely such limited overlap.
To address this, we use the estimated depth to smoothly interpolate between the input view and the first target view, freezing the first frame for a short while until the target pose is reached, then concatenate these interpolated frames with the actual input sequence.

\begin{itemize}
    \item \textbf{Generative Camera Dolly (GCD)~\cite{vanhoorick2024gcd}:}
This model only supports inference with 14 frames at a time (both in terms of input and output video), and with 3 degrees of freedom.
It assumes a spherical coordinate system $(\phi, \theta, r)$, where the camera controls provided to the network are the relative azimuth angle $\Delta \phi$, relative elevation angle $\Delta \theta$, and relative radius $\Delta r$.
The input and target viewpoints always aim at the center of the scene.
To reduce the $6 \cdot T$-DOF \ourbench camera trajectories into the 3-DOF conditioning space of GCD, information loss is unavoidable, so we apply the following approximate projection:

\begin{enumerate}
    \item Take the forward-looking vector $f=(f_x,f_y,f_z)$ (= third column of the extrinsics matrix) and translation vector $t=(t_x,t_y,t_z)$ (= last column of the extrinsics matrix) of the camera pose of each viewpoint of either the middle or last frame (depending on the dataset) of the video.
    \item Measure the azimuth angle of each vector: $\phi = \arctan \left( \frac{f_y}{f_x} \right)$; the difference between both values is then $\Delta \phi$.
    \item Measure the elevation angle of each vector: $\theta = -\arctan \left( \frac{f_z}{\sqrt{f_x^2 + f_y^2}} \right)$; the difference between both values is then $\Delta \theta$.
    \item Measure the Euclidean distance from each camera origin to the scene origin: $r=\sqrt{t_x^2+t_y^2+t_z^2}$; the difference between both values is then $\Delta r$.
\end{enumerate}

    \vspace{0.1cm}
    \item \textbf{Trajectory Attention~\cite{xiao2025trajectory}:}
TrajectoryAttention takes a variable number of input image frames at a resolution of $1024 \times 576$. Given $N$ input images, we provide the $N$ warped images from the target views along with the first image from the source view ($N+1$ images in total). Since our trajectories are represented in metric space, we opted to use the metric version of DepthAnythingV2, unlike the non-metric model used in the original implementation. We also modified the original warping code, which only supported transformations around the source view, so that it can handle arbitrary trajectories.

    \vspace{0.1cm}
    \item \textbf{GEN3C~\cite{ren2025gen3c}:}
GEN3C supports number of frames in $120 * N + 1$ pattern; we choose 121 as it is enough to cover the length of clips in all evaluated datasets. To meet the length requirement, each input video is padded to 121 frames using the last frame, and metrics are only computed on the original leading frames from the output. Following the official inference code, the videos are first resized and predicted in $1280 \times 704$, and we resize them back to the original resolution for metrics calculation. The original implementation requires per-frame camera pose, intrinsics, and depth map estimated by choice of SLAM packages (VIPE \cite{huang2025vipe} recommended) for each video; while this is designed for arbitrary videos without 3D information, it prevents us from specifying desired camera poses and intrinsics for fair comparison with the ground truths. Therefore, we instead feed the pipeline ground truth camera poses, intrinsics, and depths maps estimated by DepthAnythingV2 as mentioned in the beginning of section. It is worth noting that VIPE's estimated depth cannot be used alone in this case, as its scale is coupled with the estimated pose and intrinsics instead of ground truth ones.

    \vspace{0.1cm}
    \item \textbf{TrajectoryCrafter~\cite{mark2025trajectorycrafter}:}
TrajCrafter supports 49-frame clips at $672 \times 384$.
The input camera is flexible.
The original implementation relies on a parameterized trajectory representation ($\theta$, $\phi$, $r$, $x$, $y$) for spherical camera motion and computes geometric warping using depth estimated by DepthCrafter~\cite{hu2025-DepthCrafter}.
While suitable for smooth parametric trajectories, this approach has limited support for arbitrary real-world camera transformations, such as those found in our benchmark.
To address this limitation, we modified the inference implementation to load pre-computed re-projected RGB frames, bypassing the original depth estimation and re-projection steps.
We apply the depth warping interpolation procedure as described above.
Binary masks are automatically computed by thresholding black pixels to identify invalid re-projection regions.
The rest of the implementation is left unchanged.

    \vspace{0.1cm}
    \item \textbf{CogNVS~\cite{chen2025cognvs}:}
Similarly to TrajectoryCrafter, CogNVS supports 49-frame sequences at a resolution of $720 \times $480. We do not perform test-time optimization and instead run the model in a zero-shot manner.
CogNVS can be combined with any depth reconstruction approach, allowing improved view synthesis through better geometric reconstruction.
To ensure consistency with other baselines that rely on off-the-shelf depth estimators, we use monocular depth estimated by DepthAnythingV2.
We apply the depth warping interpolation procedure as described above, matching the required 49-frame length.

\end{itemize}

\subsection{Comparison with ReCamMaster}

In Table~\ref{tab:rcm}, we additionally evaluate ReCamMaster~\cite{bai2025recammaster} (RCM), despite it not natively supporting arbitrary camera trajectories, following the same best-effort adaptation we apply to the other baselines.
While \method consistently outperforms RCM in terms of PSNR and LPIPS, we hypothesize that the SSIM values end up being similar (despite RCM results being qualitatively worse) because the SSIM metric, which has a somewhat small default window size of $7 \times 7$, penalizes blurry (or even completely flat) regions less harshly than wrongly placed sharp edges. Since the exact geometric scale of most realistic dynamic scenes is slightly ambiguous, but the point of the benchmark is to change the camera viewpoint dramatically, if \method places an edge (even with the correct appearance) only 4 pixels away from its ground truth location, this could significantly negatively affect the SSIM numbers.

\begin{table}[t]
  \centering
  \smaller
  \setlength{\tabcolsep}{5pt}
  \renewcommand{\arraystretch}{0.99}

  \begin{tabular}{@{}l |ccc|ccc@{}}
    \toprule
    \multirow{2}{*}{Dataset}
    & \multicolumn{3}{c|}{RCM~\cite{bai2025recammaster}}
    & \multicolumn{3}{c}{Ours (\method)}
    \\
    \cmidrule(lr){2-4}\cmidrule(lr){5-7}
    & PSNR$\uparrow$ & SSIM$\uparrow$ & LPIPS$\downarrow$
    & PSNR$\uparrow$ & SSIM$\uparrow$ & LPIPS$\downarrow$
    \\

    \midrule
    \midrule
    \multicolumn{7}{@{}l}{\textbf{Narrow (zero-shot)}} \\
    \midrule
    \midrule
    DyCheck~\cite{gao2022dynamic}
    & 10.54 & 0.220 & 0.692
    & \textbf{13.02} & \textbf{0.276} & \textbf{0.556} \\  %

    \midrule
    \midrule
    \multicolumn{7}{@{}l}{\textbf{Extreme (zero-shot)}} \\
    \midrule
    \midrule
    AssemblyHands~\cite{ohkawa:cvpr23}
    & \textbf{10.54} & \textbf{0.306} & 0.738
    & 10.39 & 0.275 & \textbf{0.711} \\  %
    EgoExo4D~\cite{egoexo4d}
    & 11.80 & 0.258 & 0.676
    & \textbf{13.02} & \textbf{0.281} & \textbf{0.578} \\  %
    Argoverse~\cite{Argoverse2}
    & 11.07 & \textbf{0.427} & 0.722
    & \textbf{11.87} & 0.388 & \textbf{0.645} \\  %
    DDAD~\cite{ddad}
    & 9.50 & \textbf{0.325} & 0.740
    & \textbf{10.63} & 0.314 & \textbf{0.568} \\  %
    DROID~\cite{droid}
    & 11.22 & 0.363 & 0.676
    & \textbf{11.98} & \textbf{0.399} & \textbf{0.603} \\  %
    \midrule
    Average
    & 10.83 & \textbf{0.336} & 0.710
    & \textbf{11.58} & 0.331 & \textbf{0.621} \\  %
    \bottomrule
  \end{tabular}

  \vspace{2mm}
  \caption{
  \textbf{Comparison with ReCamMaster}~\cite{bai2025recammaster} (RCM) on AnyViewBench.
  \method outperforms, or is competitive with, RCM on DyCheck iPhone and the zero-shot datasets across all three metrics.
  }
  \label{tab:rcm}
  \vspace{-3mm}
  
\end{table}

\section{Temporal Consistency}

To assess temporal stability independently of reconstruction accuracy, we report VBench~\cite{vbench} Temporal Flickering (TF) and Motion Smoothness (MS) in Table~\ref{tab:vbench}. All methods, including the ground truth, score within a narrow band ($\geq\!95$ out of $100$), indicating that these metrics are largely saturated in our setting. We therefore interpret this as evidence that \method is at least as temporally stable as all baselines, while producing substantially more accurate content, as measured by our reconstruction and camera-pose metrics. Indeed, methods such as GCD attain high VBench scores precisely because their outputs are blurry and near-static, underscoring that such measures should only be considered jointly with ground-truth-based ones.

\begin{table}[t]
  \centering
  \smaller
  \setlength{\tabcolsep}{6pt}
  \renewcommand{\arraystretch}{0.99}
  \begin{tabular}{@{}l|cc|cc@{}}
    \toprule
    \multirow{2}{*}{Method}
    & \multicolumn{2}{c|}{In-distribution}
    & \multicolumn{2}{c}{Zero-shot}
    \\
    \cmidrule(lr){2-3}\cmidrule(lr){4-5}
    & TF$\uparrow$ & MS$\uparrow$ & TF$\uparrow$ & MS$\uparrow$
    \\
    \midrule
    GCD~\cite{vanhoorick2024gcd}        & 98.38 & 98.93 & 98.72 & 99.11 \\
    TrajAttn~\cite{xiao2025trajectory}  & 96.86 & 97.65 & 96.52 & 97.40 \\
    GEN3C~\cite{ren2025gen3c}           & 97.48 & 98.63 & 97.89 & 99.08 \\
    TrajCrafter~\cite{mark2025trajectorycrafter} & 95.63 & 97.31 & 96.01 & 97.64 \\
    CogNVS~\cite{chen2025cognvs}        & 95.92 & 97.14 & 97.29 & 98.16 \\
    Ours (\method)                      & 97.93 & 98.83 & 98.04 & 99.01 \\
    \midrule
    Ground truth                        & 97.84 & 98.84 & 97.96 & 99.04 \\
    \bottomrule
  \end{tabular}
  \vspace{2mm}
  \caption{
  \textbf{Temporal consistency} on \ourbench, measured with VBench~\cite{vbench} Temporal Flickering (\emph{TF}) and Motion Smoothness (\emph{MS}); higher is better. %
  All methods score within a narrow band of one another and of the ground truth, indicating that these metrics are largely saturated in this regime.
  }
  \label{tab:vbench}
  \vspace{-3.5mm}
  
\end{table}

\end{document}